\pdfoutput=1

\documentclass[11pt]{article}

\usepackage[final]{acl}

\usepackage{times}
\usepackage{latexsym}

\usepackage[T1]{fontenc}

\usepackage[utf8]{inputenc}

\usepackage{microtype}

\usepackage{inconsolata}

\usepackage{calc}
\usepackage{amsmath}
\usepackage{caption}
\usepackage{multirow}
\usepackage{amsfonts}
\usepackage[font=footnotesize]{subcaption}
\usepackage{booktabs}
\usepackage{pdfpages}
\usepackage{xcolor}
\usepackage{tabularx}
\usepackage{bbm}
\usepackage{graphicx}
\usepackage{algorithm}
\usepackage{algorithmic}
\usepackage{enumitem}
\newcommand{\promptplaceholder}[1]{\texttt{\{#1\}}}

\newcommand{\upsidedownface}{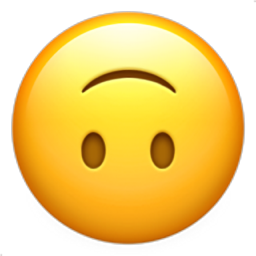}

\newcommand{\horns}{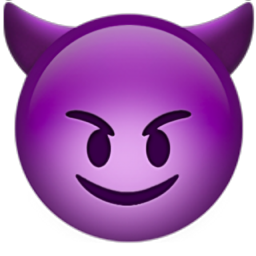}

\newcommand{\kiss}{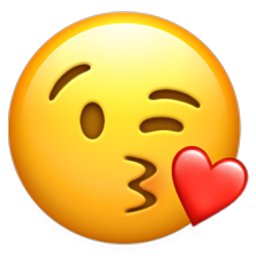}

\newcommand{\unamusedface}{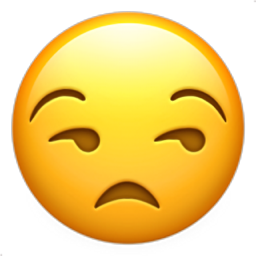}
\newcommand{\relieved}{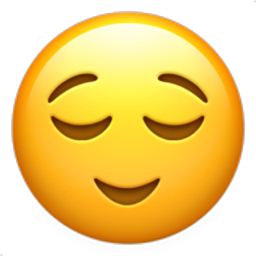}
\newcommand{\pleadingface}{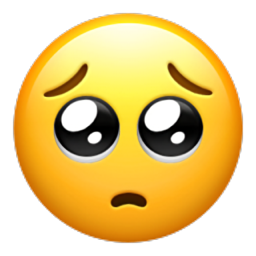}
\newcommand{\heart}{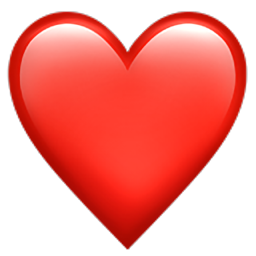}
\newcommand{\blackheart}{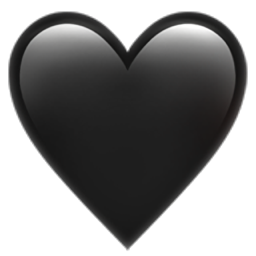}

\newcommand{\speakinghead}{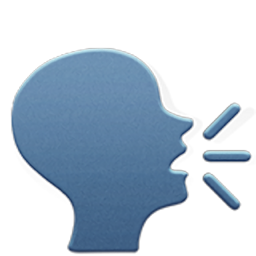}

\newcommand{\fire}{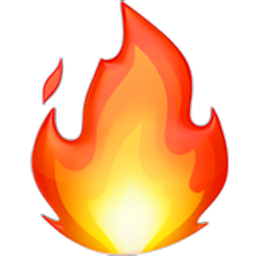}
\newcommand{\cursingface}{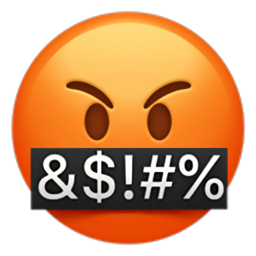}
\newcommand{\vomitingface}{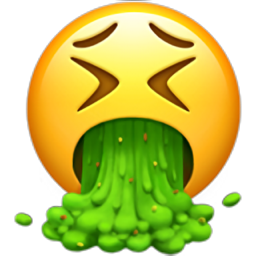}
\newcommand{\peach}{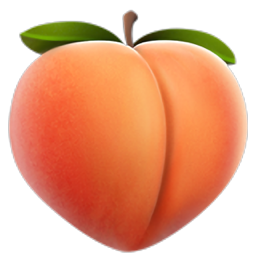}

\newcommand{\joy}{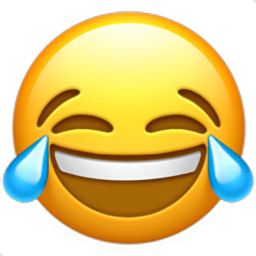}

\newcommand{\middlefinger}{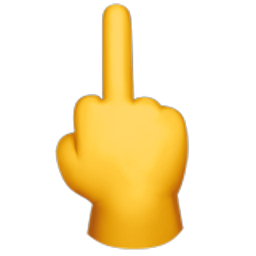}
\newcommand{\fistraised}{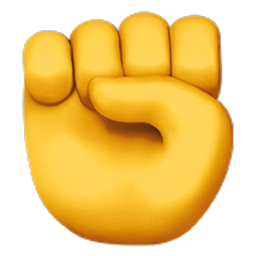}
\newcommand{\rage}{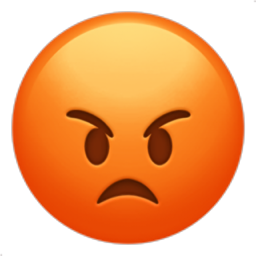}
\newcommand{\sob}{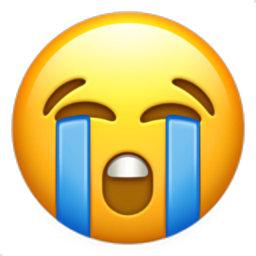}
\newcommand{\hearteyes}{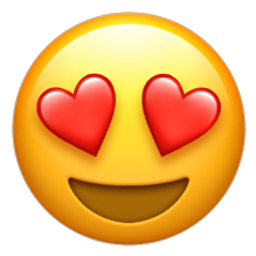}
\newcommand{\eyes}{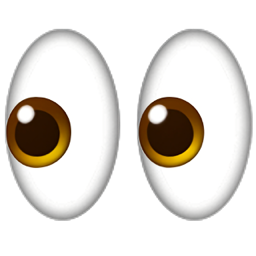}
\newcommand{\twohearts}{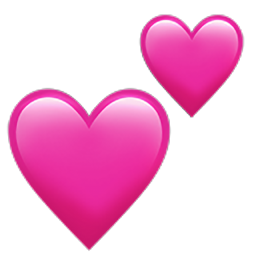}

\newcommand{\dizzy}{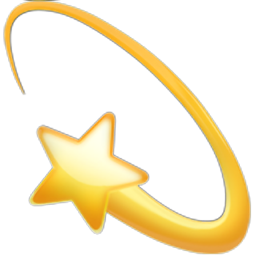}

\newcommand{\kissmark}{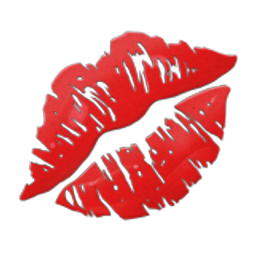}

\newcommand{\growingheart}{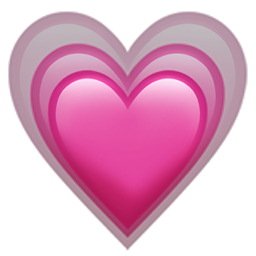}

\newcommand{\smilingfacewithsunglasses}{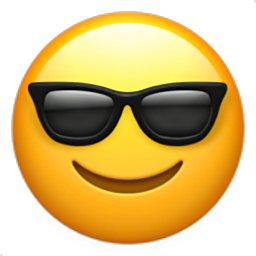}
\newcommand{\smirk}{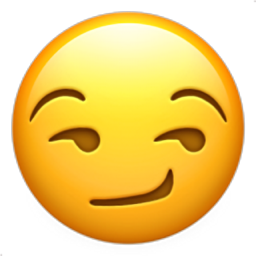}
\newcommand{\waveemoji}{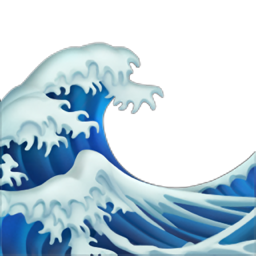}
\newcommand{\tongue}{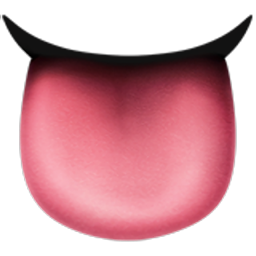}

\newcommand{\wink}{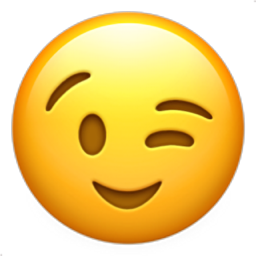}

\newcommand{\stuckouttonguewinkingface}{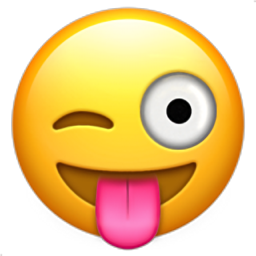}
\newcommand{\pensiveface}{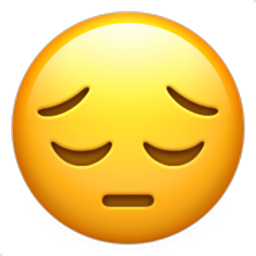}

\newcommand{\thumbsup}{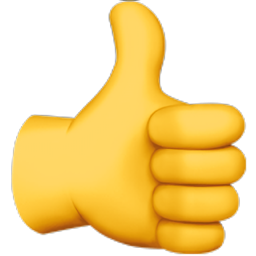}

\newcommand{\rocket}{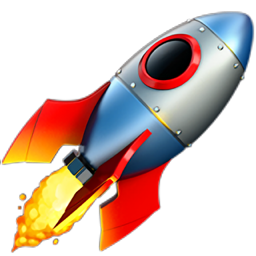}

\newcommand{\thinkingface}{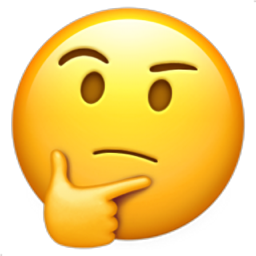}

\newcommand{\grinningface}{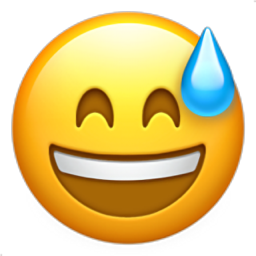}

\newcommand{\ratemoji}{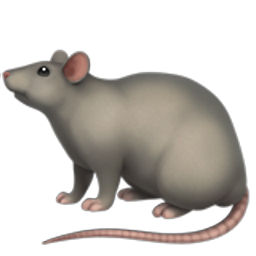}
\newcommand{\trashemoji}{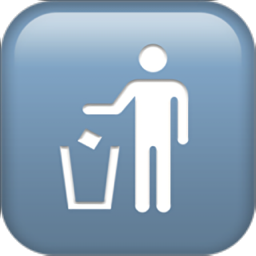}
\newcommand{\smilingfacesmilingeyes}{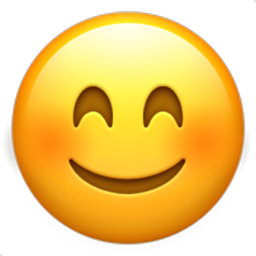}

\newcommand{\rollinglaughing}{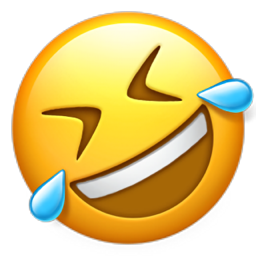}
\newcommand{\smilingfacewithhearts}{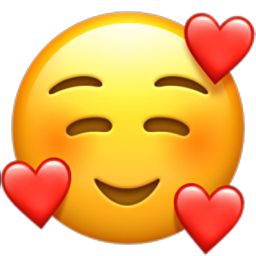}

\newcommand{\facesavoringfood}{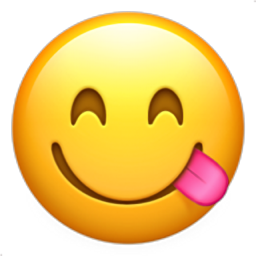}

\newcommand{\droolingface}{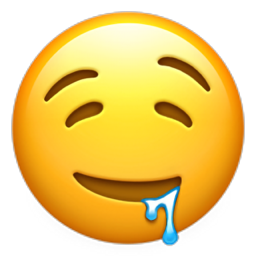}

\newcommand{\openhands}{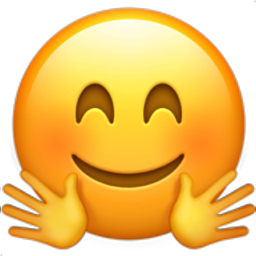}

\newcommand{\goldstar}{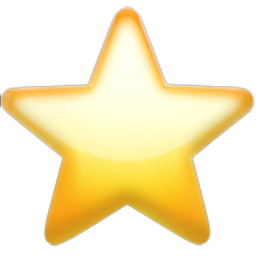}

\newcommand{\thumbsdown}{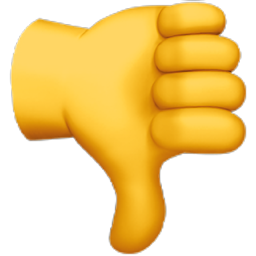}

\newcommand{\wearyface}{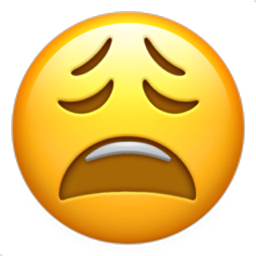}
\newcommand{\facewithsteam}{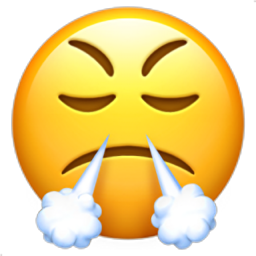}
\newcommand{\cryingface}{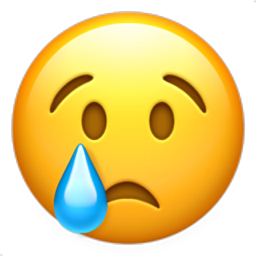}

\newcommand{\hundredpoints}{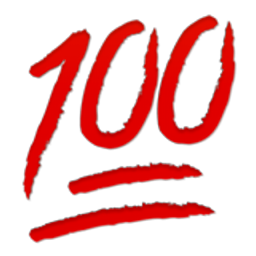}
\newcommand{\facewithrollingeyes}{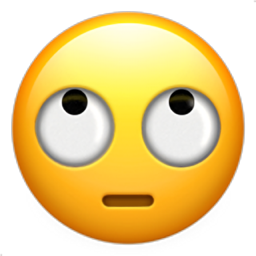}
\newcommand{\blueheart}{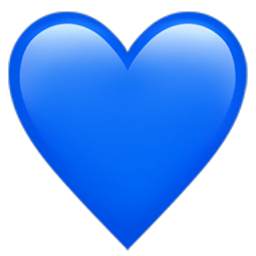}

\newcommand{\flushedface}{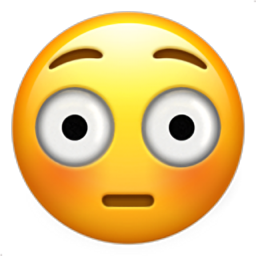}

\newcommand{\droplets}{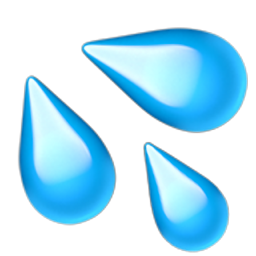}
\newcommand{\woozyface}{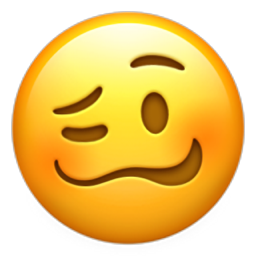}
\newcommand{\angryface}{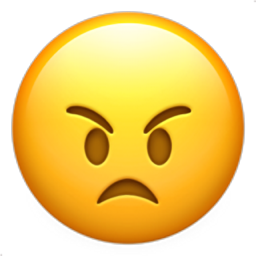}
\newcommand{\sleepingface}{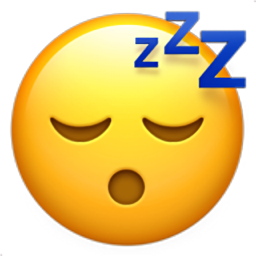}

\newcommand{\facewithopenmouth}{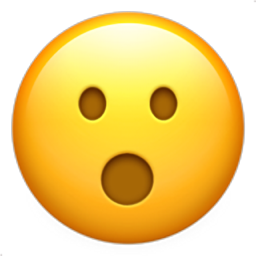}

\newcommand{\hotface}{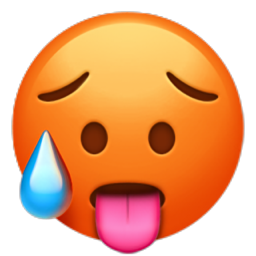}

\newcommand{\neutralface}{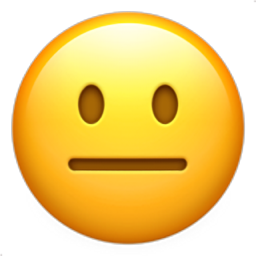}
\newcommand{\expressionlessface}{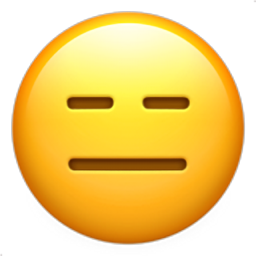}

\newcommand{\grinningsquintingface}{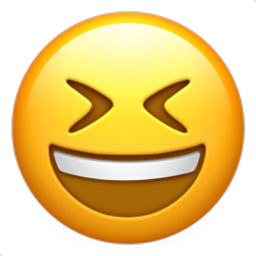}
\newcommand{\lyingface}{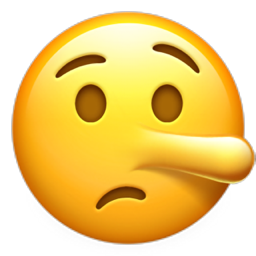}
\newcommand{\prohibitedemoji}{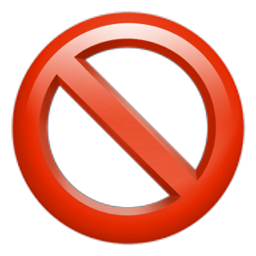}
\newcommand{\mouseface}{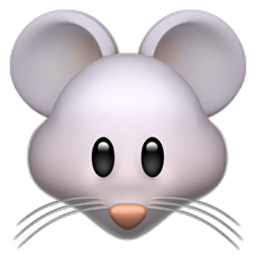}

\newcommand{\skull}{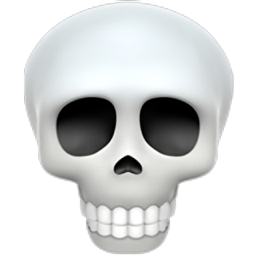}

\newcommand{\cherryblossom}{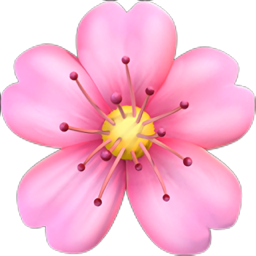}

\newcommand{\usaflag}{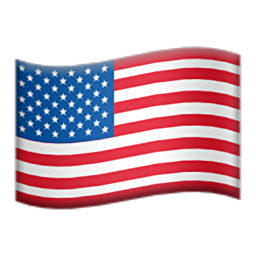}

\newcommand{\confusedface}{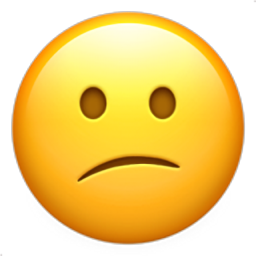}

\newcommand{\clownface}{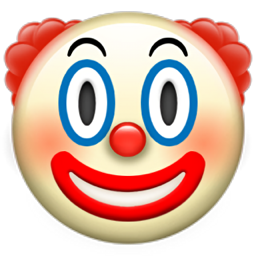}

\newcommand{\poop}{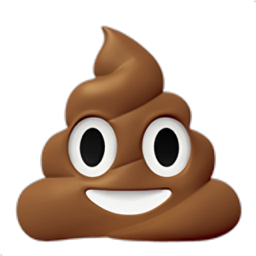}
\newcommand{\nauseated}{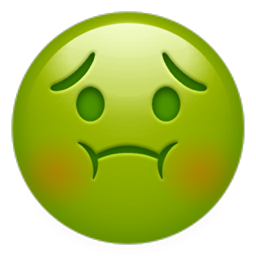}
\newcommand{\eggplant}{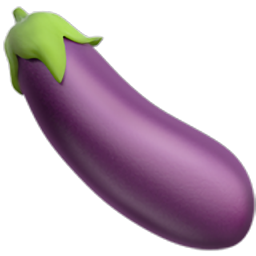}

\usepackage[export]{adjustbox}
\newcommand{\MyEmoji}[1]{\includegraphics[width=1em,valign=t]{#1}}

\title{The Hidden Language of Harm: Examining the Role of Emojis in Harmful Online Communication and Content Moderation}

\author{Yuhang Zhou \hspace{0.4cm} Yimin Xiao \hspace{0.4cm} Wei Ai \hspace{0.4cm} Ge Gao \\
University of Maryland, College Park \\
  \texttt{\{tonyzhou, yxiao, aiwei, gegao\}@umd.edu}\\
}

\begin{document}
\maketitle
\begin{abstract}
Social media platforms have become central to modern communication, yet they also harbor offensive content that challenges platform safety and inclusivity. While prior research has primarily focused on textual indicators of offense, the role of emojis, ubiquitous visual elements in online discourse, remains underexplored. Emojis, despite being rarely offensive in isolation, can acquire harmful meanings through symbolic associations, sarcasm, and contextual misuse. In this work, we systematically examine emoji contributions to offensive Twitter messages, analyzing their distribution across offense categories and how users exploit emoji ambiguity. To address this, we propose an LLM-powered, multi-step moderation pipeline that selectively replaces harmful emojis while preserving the tweet's semantic intent. Human evaluations confirm our approach effectively reduces perceived offensiveness without sacrificing meaning. Our analysis also reveals heterogeneous effects across offense types, offering nuanced insights for online communication and emoji moderation.
\end{abstract}

\section{Introduction}
\label{sec:intro}

Social media platforms host an incredibly diverse range of content, which is central to how people communicate online. However, due to varying degrees of content moderation and censorship policies, platforms like Twitter often become repositories for offensive language, threatening the cohesion and safety of online communities \cite{davidson2019racial}.
When analyzing offensive tweets, most research has focused on textual elements—explicit slurs, abusive phrases, or implicit language that reflects social biases \cite{caselli2020feel, zampieri2019predicting}. In response, scholars have developed various approaches, many leveraging Large Language Models (LLMs), to detect offensive content across cultural and linguistic contexts \cite{zhou2023cross, deng2022cold}.
Yet, despite these efforts, one critical aspect of online communication has been largely overlooked: the role of emojis in conveying or intensifying offensive messages.

Emojis, as visual symbols, are embedded in the context of communication and carry more complex semantics than individual words. 
On the one hand, unlike word tokens, very few emojis directly convey offensive meanings: exceptions include emojis like \MyEmoji{\middlefinger} (middle finger) and \MyEmoji{\poop} (pile of poop), as emojis are generally not designed with the intent to offend. 
On the other hand, the widespread use of emojis leads to varying interpretations based on their visual appearance. Emojis, with their symbolic representation of objects or ideas through similar shapes, can convey offensive meanings. For example, users often use the \MyEmoji{\peach} (peach) emoji to symbolize buttocks and the \MyEmoji{\droplets} (droplets) emoji to symbolize sperm.
Moreover, for sentiment-related emojis, one of the key characteristics of emojis is their ability to express irony or sarcasm \cite{hu2017spice}. Emojis such as \MyEmoji{\upsidedownface} (upside-down face) and \MyEmoji{\rollinglaughing} (rolling on the floor laughing) are often used to intensify offense by conveying a sarcastic tone. Even emojis typically associated with positive sentiment, such as \MyEmoji{\hearteyes} (smiling face with heart-eyes), can take on an offensive meaning when used in inappropriate contexts, such as sexual harassment. 

Given the subtle yet potent ways in which emojis contribute to offensive communication, it is essential to systematically examine their roles and functions within online discourse.
In this work, we take a first step toward that goal. We begin by identifying emojis frequently found in offensive tweets and analyzing how they relate to different types of offensive content. To deepen our understanding, we classify offensive tweets by category (e.g., personal attacks, racial slurs) and investigate which emojis are commonly used within each category.

While content moderation has traditionally focused on text \cite{zampieri2019predicting, pitsilis2018effective, husain2021survey}, we argue that emojis present a common jailbreaking way that users exploit, either deliberately or unintentionally, to convey offensive meaning through stereotypical associations.
To address this, we design a multi-step moderation pipeline powered by large language models. Rather than rewriting entire tweets, a method that risks eliminating linguistic nuance and cultural expression, we target only the emojis likely to cause harm.
The pipeline is implemented to identify emojis that are either directly related to offensive content or have the potential to evoke offense, and recommend emoji surrogates that preserve the tweet's semantics. Human evaluations show that this pipeline effectively reduces offensiveness while maintaining the tweet’s meaning. We also analyze its heterogeneous effects across different tweet types and examine the relationship between emoji functionality and offensiveness.

We summarize our contributions as follows:
\begin{itemize}[leftmargin=*]
    \item We explore the relationship between emojis and offensive content in online communication, examining the roles emojis play under different offensive types. 
    \item We design and implement a multi-step LLM pipeline to better moderate offensive emojis in tweets and recommend emoji surrogates. 
    \item We conduct a human evaluation to demonstrate the effectiveness of our pipeline and analyze its heterogeneous effects across various offensive types.
\end{itemize}

\section{Related Work}
\label{sec:related}

Our work is based on two lines of existing work: emoji functionality and offensive content detection.

\paragraph{Emoji Functionality and Interpretation}
Emojis, as prevalent visual elements, have attracted the interest of researchers. The semantics embedded in emojis extend beyond a single word token, giving their various functionalities such as expressing sentiments and irony, softening tones, and enhancing communication \cite{ai2017untangling, ge2019identity, hu2017spice, miller2016blissfully, cramer2016emojifunction}. 
The rich meanings and diverse functionalities of emojis make them useful in various tasks, including sentiment analysis, predicting user behavior, and increasing communication on social media \cite{Felbo_2017, chen2018gender, chen2019emojipowered, zhou2023emoji, zhou2022emoji}.
However, aside from emojis' use in expressing irony or sarcasm, most studies focus on the positive effects of emojis, overlooking the fact that negative sentiments or visual symbols in emojis can also be used to offend others. In this work, we focus on the patterns of emoji usage in offensive tweets and explore how users employ emojis to convey offensive content.

\paragraph{Offensive Content Detection}

The prevalence of social networks has encouraged users to develop more flexible forms of offensive behavior. In response, researchers have examined patterns of offense in online communication and developed various methods to detect and mitigate offensive content in text \cite{davidson2019racial, davidson2017automated, pitsilis2018effective, poletto2021resources}.
With the advancement of LLMs, there is increasing interest in leveraging them for the effective detection of hate speech and other hidden performance bias \cite{huang2023chatgpt, li2023hot, zhu2023can, zhou2025mergeme}. Furthermore, due to their text generation capabilities, some studies have used LLMs to augment collected datasets, thus enhancing the robustness of hate speech detection models \cite{xiao2024toxicloakcn, nghiem2024hatecot}.

Beyond general offense, researchers have also explored offenses of different types, including sexual \cite{vandenbosch2015relationship}, racial \cite{davidson2019racial}, and violent offenses \cite{zhong2019detecting}, as each offense type tends to target different groups and exhibit specific linguistic patterns.
Given the strong link between offense and culture, researchers have also explored offensive content across multiple languages, including English \cite{pitsilis2018effective}, Chinese \cite{deng2022beike}, Arabic \cite{husain2021survey}, and French \cite{battistelli2020building}. As a result, more advanced models and increasingly diverse datasets have emerged to enhance the detection of offensive content. In this work, we contribute understanding of emoji use patterns in various types of offensive content, and leverage LLMs to enhance the detection and mitigation of emoji-relevant offensiveness in online content.
\section{Understanding Emojis in Offensive Contexts}
\label{sec:emoji_offensive}
We begin our exploration of emoji functionality in offensive contexts by analyzing their roles and distributions in offensive tweets. To identify offensive content, we collected tweets from January 1 to December 31, 2019, via the Twitter API\footnote{\url{https://developer.twitter.com/en/docs/twitter-api}}. We then used a pre-trained RoBERTa model (cardiffnlp/twitter-roberta-base-offensive \footnote{https://huggingface.co/cardiffnlp/twitter-roberta-base-offensive}) to filter candidate tweets with offensive content \cite{liu2019roberta, barbieri2020tweeteval}. Tweets with a predicted probability greater than 0.5 were selected, and GPT-4 (ChatGPT) was used to further label whether the tweet contained offensive content \cite{openai2023gpt4}. This process resulted in 9,285 annotated offensive tweets. To ensure precise annotation, we define offensive content as posts containing unacceptable language (profanity) or targeted offenses, whether direct or veiled, including insults, threats, profane language, or swear words, following the definition used in previous work \cite{poletto2021resources, zampieri2019predicting}.

\subsection{Emoji Role in Offensive Tweets}
\label{sec:emoji_role}
To understand how emojis function within offensive tweets, we developed a taxonomy grounded in literature on emoji functions (Section \ref{sec:related}) and their interaction with offensive language. We categorize emoji roles based on their relationship to the offense: 
\begin{itemize}[leftmargin=*]
\item \textbf{Offensive in itself.} The emoji alone constitutes an offense, such as \MyEmoji{\middlefinger} (middle finger). 

\item \textbf{Intensify offense.} Emojis can enhance the intensity of an offensive tweet by expressing irony or sarcasm \cite{weissman2018strong}, thereby amplifying its offensive nature. 

\item \textbf{Mitigate offense.} Emojis can also soften or adjust the tone of a tweet, reducing its offensive impact \cite{cramer2016emojifunction, ge2019identity}. 

\item \textbf{Unrelated to offense.} The emoji is not directly connected to the offensive content of the tweet. 

\end{itemize}

\begin{figure}[tbp]
     \centering
     \includegraphics[width=\linewidth]{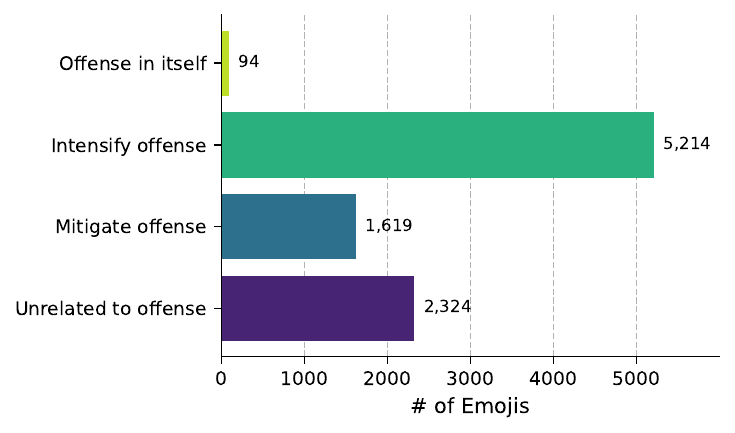}
     \caption{\label{fig:emoji_func_dist} Distribution of emoji role in offensive tweets}
\end{figure}

Based on the proposed taxonomy, we use GPT-4 to annotate the role of all emojis present in the collected offensive tweets. The distribution of emoji roles is illustrated in Figure \ref{fig:emoji_func_dist}.

The distribution (Figure \ref{fig:emoji_func_dist}) reveals that most annotated emojis intensify, mitigate, or are unrelated to the offense, with few being offensive in themselves, as expected given that few emojis carry inherently offensive meanings. The prevalence of intensification suggests a notable link between emojis and offensive expression.

Table \ref{table:emoji_func_example} lists the top 10 emojis for each role. Emojis categorized as `Offensive in itself` (e.g., \MyEmoji{\poop}, \MyEmoji{\middlefinger}, \MyEmoji{\peach}) show no overlap with other categories and are often used for direct insults. Conversely, significant overlap exists among the top emojis for the other three roles. Emojis like \MyEmoji{\joy} and \MyEmoji{\sob} appear frequently across these categories, highlighting their context-dependent functions. Notably, even positive emojis like \MyEmoji{\hearteyes} can intensify offense, particularly in contexts like sexual harassment.

\begin{table}[tbp]
    \small
    \centering
    \begin{tabular}{ l  l }
        \toprule
        Role & Top 10 Frequent Emojis  \\
        \midrule
        Offensive in itself &  \MyEmoji{\clownface}\MyEmoji{\poop}\MyEmoji{\nauseated}\MyEmoji{\cursingface}\MyEmoji{\middlefinger}\MyEmoji{\horns}\MyEmoji{\eggplant}\MyEmoji{\vomitingface}\MyEmoji{\peach}\MyEmoji{\rage} \\
        \midrule
        Intensify offense &  \MyEmoji{\joy}\MyEmoji{\sob}\MyEmoji{\rollinglaughing}\MyEmoji{\facewithrollingeyes}\MyEmoji{\skull}\MyEmoji{\woozyface}\MyEmoji{\wearyface}\MyEmoji{\unamusedface}\MyEmoji{\hundredpoints}\MyEmoji{\hearteyes}  \\
        \midrule
        Mitigate offense & \MyEmoji{\joy}\MyEmoji{\sob}\MyEmoji{\pleadingface}\MyEmoji{\pensiveface}\MyEmoji{\rollinglaughing}\MyEmoji{\grinningface}\MyEmoji{\wearyface}\MyEmoji{\woozyface}\MyEmoji{\heart}\MyEmoji{\cryingface}  \\
        \midrule
        Unrelated to offense &  \MyEmoji{\joy}\MyEmoji{\sob}\MyEmoji{\hearteyes}\MyEmoji{\rollinglaughing}\MyEmoji{\thinkingface}\MyEmoji{\fire}\MyEmoji{\wearyface}\MyEmoji{\woozyface}\MyEmoji{\heart}\MyEmoji{\eyes}  \\
        \bottomrule
    \end{tabular}
    \caption{Top-10 emojis under each emoji role in offensive tweets}
    \label{table:emoji_func_example}
\end{table}



Given that specific emojis (e.g., \MyEmoji{\peach}) seem linked to particular offensive themes, we next apply topic modeling to further explore emoji usage across different types of offensive content.

\subsection{Emojis Associated with Different Offensive Topics}
\label{sec:emoji_topics}
To better understand the offensive context in which each emoji appears, we identify the latent offensive types embedded in each tweet and explore the emojis associated with each specific type. To summarize the offensive types across tweets, we first clustered tweets into distinct topics using unsupervised topic modeling (BerTopic \cite{grootendorst2022bertopic}). We then extracted representative words (ranked by tf-idf) for each topic and used GPT-4 to generate topic descriptions \cite{aizawa2003information}. We set a minimum threshold of 20 documents per cluster for our dataset, resulting in the identification of 14 distinct topics. Using the topic descriptions and representative keywords, we employed GPT-4 to summarize the offensive types and align each topic with its corresponding offense category. The types and their associated topic descriptions are presented in Table \ref{table:offense_topic} in Appendix \ref{sec:appendix_supplementary_result}. Note that our derived categories align well with established offense types like sexual, personal, and violent offenses from related work (Section \ref{sec:related}), validating our GPT-4-assisted thematic grouping.

Based on the associated topics under each type, we further summarize the taxonomy of each offensive type, as outlined below:
\begin{itemize}[leftmargin=*]
    \item \textbf{Sexual Content and Gender Issues}: This offensive type includes sexual harassment, gender discrimination, body shaming, and objectification. Gender-based insults and derogation also fall into this category.
    \item \textbf{Personal Attacks and Disrespect}: This includes direct insults, disrespect, or derogation targeting individuals based on personal characteristics.
    \item \textbf{Racial and Ethnic Offense}: This includes racial slurs, ethnic stereotyping, and various forms of discrimination and prejudice based on race or ethnicity.
    \item \textbf{Political and Social Issues}: This includes political attacks, social discrimination, harassment, and aggression against individuals or groups over their political views.
    \item \textbf{Violence and Abuse}: This includes topics related to physical or verbal abuse and violence. This can be related to threats, aggressive behaviors, and other forms of violence as forms of offensive content.
\end{itemize}

We aggregate the emojis within each topic and use the matching relationship between topics and offensive types to assign emojis to each offensive type. In Table \ref{table:emoji_type_example}, we present the top 10 most frequent emojis for each offensive type. We note that for the type of ``political and social issues'' of offense, only 8 emojis are present.

\begin{table}[tbp]
    \small
    \centering
    \begin{tabular}{ l  l }
        \toprule
        Offense Type & Top 10 Frequent Emojis  \\
        \midrule
        Sexual Content &  \MyEmoji{\droplets}\MyEmoji{\eggplant}\MyEmoji{\tongue}\MyEmoji{\hearteyes}\MyEmoji{\peach}\MyEmoji{\horns}\MyEmoji{\facesavoringfood}\MyEmoji{\droolingface}\MyEmoji{\fire}\MyEmoji{\kissmark} \\
        \midrule
        Personal Attacks and Disrespect & \MyEmoji{\poop}\MyEmoji{\joy}\MyEmoji{\clownface}\MyEmoji{\facewithrollingeyes}\MyEmoji{\cursingface}\MyEmoji{\vomitingface}\MyEmoji{\ratemoji}\MyEmoji{\rollinglaughing}\MyEmoji{\middlefinger}\MyEmoji{\rage} \\
        \midrule
        Racial and Ethnic Offense & \MyEmoji{\joy}\MyEmoji{\hundredpoints}\MyEmoji{\sob}\MyEmoji{\woozyface}\MyEmoji{\rollinglaughing}\MyEmoji{\skull}\MyEmoji{\facewithrollingeyes}\MyEmoji{\speakinghead}\MyEmoji{\trashemoji}\MyEmoji{\smirk}  \\
        \midrule
        Political and Social Issues & \MyEmoji{\ratemoji}\MyEmoji{\poop}\MyEmoji{\cursingface}\MyEmoji{\rage}\MyEmoji{\clownface}\MyEmoji{\joy}\MyEmoji{\middlefinger}\MyEmoji{\usaflag}  \\
        \midrule
        Violence and Abuse & \MyEmoji{\cursingface}\MyEmoji{\rage}\MyEmoji{\facewithsteam}\MyEmoji{\joy}\MyEmoji{\angryface}\MyEmoji{\speakinghead}\MyEmoji{\hundredpoints}\MyEmoji{\skull}\MyEmoji{\sob}\MyEmoji{\unamusedface}   \\
        \bottomrule
    \end{tabular}
    \caption{Top-10 emojis under different offense types. Note that there are 8 emojis for the offense type: political and social issues, in our dataset.}
    \label{table:emoji_type_example}
\end{table}

From Table \ref{table:emoji_type_example}, we observe that different offensive types are associated with distinct sets of frequently used emojis. The emojis used often reflect the offensive nature of the tweet. For tweets classified as ``Sexual Content,'' we find that users frequently employ emojis such as \MyEmoji{\droplets} (droplets), \MyEmoji{\eggplant} (eggplant), and \MyEmoji{\tongue} (tongue) to symbolize body parts. Emojis like \MyEmoji{\hearteyes} (smiling face with heart-eyes) and \MyEmoji{\kissmark} (kiss), which typically convey positive sentiment, are used in these contexts to amplify the offensiveness when combined with sexual content. 
For the ``Personal Attacks,'' ``Racial Offense,'' and ``Political Issues'' categories, item-related emojis such as \MyEmoji{\poop} (pile of poo), \MyEmoji{\trashemoji} (trash), and \MyEmoji{\ratemoji} (rat) are commonly used to dehumanize the target and intensify the offensive content.
Moreover, for the "Violence and Abuse" category, the most frequent emojis, such as \MyEmoji{\rage} (rage) and \MyEmoji{\cursingface} (cursing face), reflect users' aggressive emotions and sentiments.
These findings demonstrate that specific emojis are closely related to the offensive context of the tweet, amplifying the underlying harmful content.

Now that we have explored the prevalent offensive types and their associated emojis, we are also interested in understanding whether these emojis are predominantly used in offensive content or appear more frequently in unoffensive content. In the next section, we will address this question by quantifying the distribution of emojis across unoffensive and offensive tweets.

\subsection{Emoji Distribution: Usage in Offensive vs. Non-Offensive Tweets}


To quantify this distribution and determine whether the emojis listed in Table \ref{table:emoji_func_example} are predominantly associated with offensive content, we analyzed usage patterns: For each target emoji, we randomly collected a sample of 1,000 tweets containing it and utilized GPT-4o to annotate each tweet for offensive content. Based on these annotations, we calculated the proportion of offensive tweets within the sample for that specific emoji.

This percentage reflects how frequently an emoji co-occurred with offensive content within our collected samples. We then categorized the emojis based on this calculated offensive content rate:
\begin{itemize}[leftmargin=*]
    \item \textbf{High Frequency:} Emojis where over 30\% of the tweets in their respective sample were offensive.
    \item \textbf{Moderate Frequency:} Emojis where 20\% to 30\% of the tweets in their respective sample were offensive.
    \item \textbf{Low Frequency:} Emojis where less than 20\% of tweets in their respective sample were offensive.
\end{itemize}

Table \ref{table:emoji_dist_example} presents the emojis categorized according to their association level with offensive content based on this analysis.

\begin{table}[tbp]
    \small
    \centering
    \begin{tabular}{ l  p{3cm} }
        \toprule
        Frequency in offensive tweets & Emojis  \\
        \midrule
        High frequency & 
        \MyEmoji{\middlefinger}\MyEmoji{\poop}\MyEmoji{\eggplant}\MyEmoji{\ratemoji}\MyEmoji{\cursingface}\MyEmoji{\trashemoji}\MyEmoji{\vomitingface}\MyEmoji{\tongue}\MyEmoji{\droplets}\MyEmoji{\clownface} \\
        \midrule
        Moderate frequency & 
        \MyEmoji{\nauseated}\MyEmoji{\angryface}\MyEmoji{\peach}\MyEmoji{\rage}\MyEmoji{\smirk}\MyEmoji{\horns}\MyEmoji{\facewithsteam}\MyEmoji{\skull}\MyEmoji{\speakinghead} \\
        \midrule
        Low frequency & 
        \MyEmoji{\facewithrollingeyes}\MyEmoji{\woozyface}\MyEmoji{\rollinglaughing}\MyEmoji{\joy}\MyEmoji{\droolingface}\MyEmoji{\hundredpoints}\MyEmoji{\facesavoringfood}\MyEmoji{\hotface}\MyEmoji{\kissmark}\MyEmoji{\smirk} \\
        & \MyEmoji{\sob}\MyEmoji{\wearyface}\MyEmoji{\thinkingface}\MyEmoji{\fire}\MyEmoji{\kiss}\MyEmoji{\hearteyes} \\
        \bottomrule
    \end{tabular}
    \caption{Categorization of Emojis Based on Frequency in Offensive vs. Non-Offensive Tweets.}
    \label{table:emoji_dist_example}
\end{table}

Table \ref{table:emoji_dist_example} indicates that most emojis appearing in offensive tweets with a high frequency are item-related, used to dehumanize others or suggest body parts. For moderately frequent emojis, we find that they predominantly convey negative sentiment, such as \MyEmoji{\facewithsteam} and \MyEmoji{\rage}. In contrast, emojis with positive or neutral sentiment are used less frequently in offensive tweets. This suggests that offensive tweets are more likely to be accompanied by negative or dehumanizing emojis.
\section{Reduce Offense with Language Models}

The analysis so far demonstrates the role of emojis in offensive tweets, the types of offenses and their associated emojis, and the distribution of emojis in offensive versus non-offensive tweets. These findings emphasize that emojis play a significant role in conveying offense, with their function closely tied to the context of the tweet. 
Our next question is how to reduce offensive content inside the tweets. The most straightforward approach would be to completely rewrite the tweet, including emojis and other tokens. However, we argue that this method, such as rewriting with tools such as ChatGPT \cite{openai2023gpt4}, would lead to a significant reduction in diversity, both lexical and content-related, as suggested in previous work \cite{padmakumar2023does, guo2023curious}. Maintaining diversity is crucial for ensuring the richness of communication and expression on social media platforms. Instead, by selectively replacing offensive emojis with ones that mitigate offense, we can reduce the overall level of offense while still preserving the diversity of expression on social networks.

\subsection{Task Setup}

Our goal for emoji replacement is to identify emojis that can mitigate offense while preserving the overall semantic meaning of the tweet. To achieve this, we have designed a multi-step pipeline that leverages LLMs to iteratively select appropriate emoji surrogates. This process is guided by our findings from Section \ref{sec:emoji_offensive}, which we incorporate into the prompts to ensure that the replacements align with the original context and sentiment of the tweet. 
We consider the input to be a user's tweet, which may or may not contain offensive content, and the output of our pipeline is a revised tweet with recommended emoji surrogates, along with a justification for each recommendation.

\subsection{Multi-step Pipeline to Identify Emoji Surrogate}
\label{sec:llm_reduce}
\begin{figure}[tbp]
     \centering
     \includegraphics[width=\linewidth]{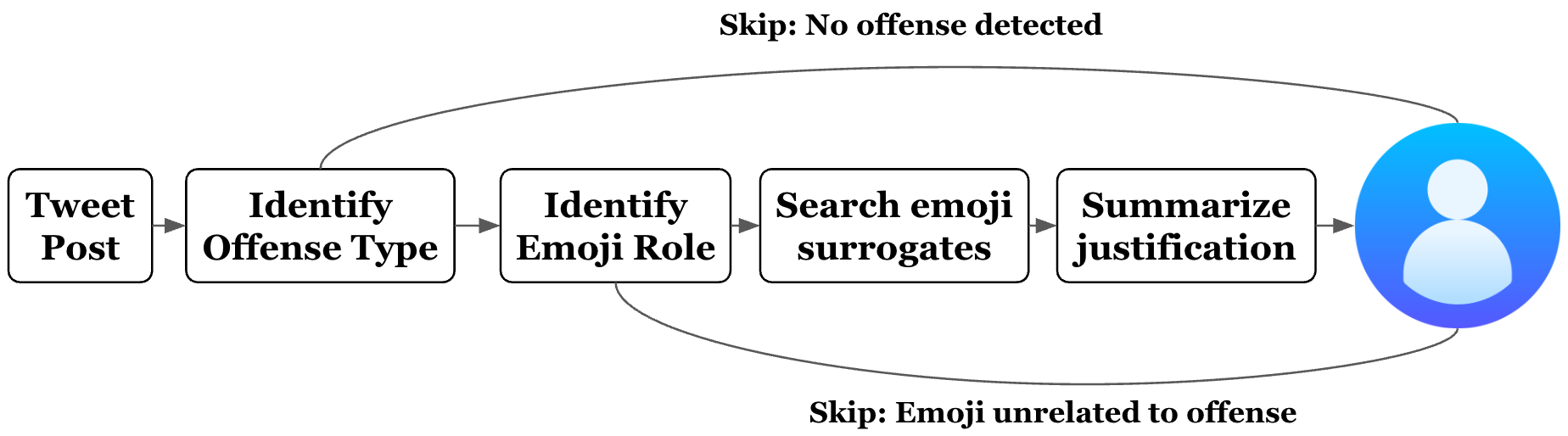}
     \caption{\label{fig:emoji_multi_step} Multi-step pipeline to suggest the emoji surrogates for selected tweets.}
\end{figure}

Our pipeline consists of four steps. Initially, we classify whether the tweet contains offensive content. If offensive content is detected, we then categorize the roles of each emoji into one of four categories, as described in Section \ref{sec:emoji_role}. Once we identify emojis that either serve as the offense themselves or intensify the offensive tone, we leverage LLMs to suggest appropriate emoji surrogates. We present the visualization in Figure \ref{fig:emoji_multi_step}. Below, we provide a detailed explanation of each step and the details of prompts are shown in Appendix \ref{sec:appendix_direct_prompt}.

\paragraph{Offensive Content Classification}
Our first step is to determine whether the tweet contains offensive content. If offensive content is identified, we prompt LLMs to classify it into one of the predefined offensive types (sexual, personal attacks, racial offense, political issues, or violence). To guide the model, we include the taxonomy of offensive types within the prompt, along with two demonstrations: one featuring an offensive tweet and the other a non-offensive tweet, to serve as examples for accurate classification. For non-offensive tweets, we leave them as is, while offensive tweets are passed to the next stage.

\paragraph{Determining the Role of Each Emoji}
In the second step, we identify the role of each emoji. We provide the LLM with the identified offense type (from step 1), our four-dimensional emoji role taxonomy (offense itself, intensifying, mitigating, unrelated), and exemplars for each role. Crucially, we incorporate findings from our analysis (Section \ref{sec:emoji_offensive}), such as common emojis for each offense type and role, and their general offensive frequency, into the prompt to guide the LLM towards more contextually accurate role identification.

\paragraph{Recommending Emoji Surrogates}
In this step, we query LLMs to recommend emoji surrogates for those identified as intensifying offense or serving as offense itself. We ask the LLMs to suggest emoji replacements that remain consistent with the original content and sentiment of the tweet. To improve the quality of the recommendations, we include two demonstrations within the prompt.

\paragraph{Summarize the justification}
Finally, we aggregate all the justifications from the previous stages and feed them back to the LLM, asking it to summarize the reasoning into a paragraph. The output presented to the users will include the revised tweet with recommended emoji surrogates, along with a summarized justification explaining how the emoji substitutions reduce the offense level.


\subsection{Experiment and Setup}
For the experiment, we use GPT-4o as the LLM to recommend emoji surrogates and run the multi-step pipeline on our collected 9,285 offensive tweets (Section \ref{sec:emoji_offensive}). To demonstrate effectiveness, we compare it with a \textbf{direct prompting} baseline, where the LLM is simply asked to replace emojis in offensive tweets to mitigate offense while maintaining tone (prompt details are shown in Appendix \ref{sec:appendix_direct_prompt}).

After running our proposed pipeline, we generated emoji surrogates for a total of 7,142 tweets. For the remaining offensive tweets, no emojis were identified as playing a role in intensifying offense or directly representing offensive content.

\subsection{Qualitative Evaluation}

\paragraph{Emoji Distribution after Substitution}
We first examine whether offensive emojis were effectively eliminated by comparing the emoji distributions for each offensive type before and after running our pipeline and the baseline. Our analysis shows the multi-step pipeline successfully eliminates most item-based emojis frequently used for offense (e.g., \MyEmoji{\poop}, \MyEmoji{\droplets}, \MyEmoji{\eggplant}) and reduces the frequency of negative sentiment emojis (\MyEmoji{\rage}, \MyEmoji{\cursingface}) and those with sarcastic tones (\MyEmoji{\facewithrollingeyes}). In contrast, the direct prompting method retained many problematic emojis (e.g., \MyEmoji{\droplets}, \MyEmoji{\peach} in sexual content). This suggests our multi-step approach, informed by prior analysis, is more effective at filtering harmful emojis. (The full emoji comparison in Table \ref{table:emoji_comparison} in Appendix \ref{sec:appendix_emoji_dist}).

\paragraph{Case Studies}
To illustrate the pipeline's effectiveness on individual tweets, we present several case studies covering different offensive types. These examples showcase how the pipeline identifies the role of emojis in context (e.g., symbolic meaning, intensification) and recommends appropriate, less offensive surrogates while providing step-by-step justifications. These detailed examples and the LLM's reasoning are provided in Appendix \ref{sec:appendix_case_study}.

\subsection{Human Evaluation}

\paragraph{Evaluation Design}
The goal of our pipeline is to reduce offensiveness while preserving semantics. A straightforward evaluation approach is to present the original and processed tweets side by side and ask users to assess whether the pipeline effectively reduces offensive content and whether semantic meaning is preserved. However, this method may introduce cognitive bias, as users might be inclined to perceive the two versions as inherently similar \cite{haselton2015evolution}.

To mitigate this bias, we conduct a within-subject user experiment in which annotators evaluate each tweet independently by answering a series of questions related to both semantics and offensiveness. Each annotator assesses 60 tweets presented in a randomized order, consisting of 20 original tweets, 20 versions of these tweets processed by our pipeline, and 20 versions processed by the baseline method. We then compare the differences in responses to the tweets before and after processing to assess the pipeline’s impact.
Ideally, annotators should perceive the rewritten tweets as less offensive while maintaining consistent responses to semantic-related questions.
\paragraph{Measures}
We collected measures that aim to assess the Offensiveness and Semantic aspect of each tweet. For Offensiveness, we firstly measured the perceived level of offensiveness of each tweet by asking  annotators to rate how offensive they find each tweet on a scale from 1 to 5, ranging from `not offensive' to `extremely offensive.' Additionally, based on the findings in Sections \ref{sec:emoji_role} and \ref{sec:emoji_topics}, we observe that emojis can enhance offensiveness by expressing irony, symbolizing body parts, or dehumanizing the target. To better interpret our pipeline's impact, we ask annotators whether the emojis in each tweet exhibit these functionalities.

We also consider the influence of emoji replacement on the semantic aspect of the tweets, given emojis' role in conversations, such as conveying sentiment or external meaning, suggested in prior research  \cite{zhou2024emojis}. For each tweet, we ask annotators to assess the sentiment, emotion arousal, whether the emojis contribute external meaning, clarity, and fluency. These annotations allow us to quantify the semantic integrity of each tweet. We present the collected measured variables as well as the metrics in Table \ref{tab:question_stats} and the questionnaire and recruitment process in Appendix \ref{sec:appendix_emoji_annotation}.

\begin{table}[tbp]
    \centering
    \footnotesize
    \begin{tabular}{@{} l p{5.5cm} @{}} 
    \toprule
    \textbf{Category} & \textbf{Measured Variables} (\textit{Scale / Type}) \\ 
    \midrule
    Offensiveness & Offensiveness Score \textit{(1-5)}, Sarcasm \textit{(\% Yes)}, Body Symbol Emojis \textit{(\% Yes)}, Dehumanizing Emojis \textit{(\% Yes)} \\
    \addlinespace 
    Semantics & Sentiment Score \textit{(1-3)}, Arousal Score \textit{(1-3)}, Emoji Extra Meaning \textit{(\% Yes)}, Clarity \textit{(\% Yes)}, Fluency \textit{(\% Yes)} \\
    \bottomrule
    \end{tabular}
    \vspace{-0.5em}
    \caption{Measured variables for tweet annotation. Detailed variable meanings are shown in Appendix \ref{sec:appendix_emoji_annotation}}
    \label{tab:question_stats}
\end{table}

\paragraph{Average Evaluation Result}

We sample 200 offensive tweets from our collected data and present annotators with 600 tweets in total: 200 original tweets, 200 tweets from our multi-step pipeline, and 200 tweets via direct prompting. Each tweet is annotated by two annotators with the predefined questions. After annotation, we compute the average scores for overall offensiveness, sentiment, and arousal, as well as the percentage of `Yes' responses for other variables. The results for the original, pipeline-processed, and direct-prompting-processed tweets are shown in Table \ref{tab:focus3_main_result}.

\begin{table}[tbp]
    \centering
    \resizebox{\linewidth}{!}{%
    \begin{tabular}{lccc}
        \toprule
        \textbf{Measured Variables} & \textbf{Original} & \textbf{Direct} & \textbf{Multi-step} \\
        \midrule
        Offensiveness Score (1-5) & $3.00$ & $2.94$ & $2.58^{*}$ \\
        Sarcastic (\% Yes) & $34.0\%$ & $38.0\%$ & $37.5\%$ \\
        Body Symbol (\% Yes) & $49.5\%$ & $51.0\%$ & $52.0\%$ \\
        Dehumanization (\% Yes) & $20.0\%$ & $12.5\%$ & $16.5\%$ \\
        \midrule
        Sentiment Score (1-3) & 1.48 & 1.59 & 1.64 \\
        Arousal Score (1-3) & 2.14 & 2.06 & 2.01 \\
        Extra Meaning (\% Yes) & 25.0\% & 24.5\% & 19.0\% \\
        Clarity (\% Yes) & 74.0\% & 70.5\% & 72.5\% \\
        Fluency (\% Yes) & 77.0\% & 76.0\% & 77.0\% \\
        \bottomrule
    \end{tabular}

    }
    \caption{Human evaluation results comparing our proposed multi-step pipeline with the direct prompting method. Our approach significantly reduces offensiveness while preserving semantics. Statistical significance is indicated as follows: *: $p<0.05$ (paired $t$-test).}
    \label{tab:focus3_main_result}
\end{table}


As shown in Table \ref{tab:focus3_main_result}, our proposed multi-step pipeline significantly reduces the offensiveness scores assigned by annotators. In terms of semantic preservation, tweets processed by our pipeline exhibit no notable changes in meaning. Compared to our pipeline, the direct prompting baseline achieves only a minor and statistically insignificant reduction in offensiveness. We suspect this is because, without prior knowledge of the relationship between offensiveness and emojis, LLMs struggle to identify suitable emoji surrogates.

\begin{table*}[tbp]
    \centering
    \resizebox{\textwidth}{!}{%
    \begin{tabular}{lcc cc cc cc cc}
        \toprule
        \textbf{Measurement Variables ($\Delta$)} & \multicolumn{2}{c}{\textbf{Personal Attacks}} & \multicolumn{2}{c}{\textbf{Political/Social}} & \multicolumn{2}{c}{\textbf{Racial/Ethnic}} & \multicolumn{2}{c}{\textbf{Sexual/Gender}} & \multicolumn{2}{c}{\textbf{Violence/Abuse}} \\
        \cmidrule(lr){2-3} \cmidrule(lr){4-5} \cmidrule(lr){6-7} \cmidrule(lr){8-9} \cmidrule(lr){10-11}
        & \textbf{Direct} & \textbf{Multi-step} & \textbf{Direct} & \textbf{Multi-step} & \textbf{Direct} & \textbf{Multi-step} & \textbf{Direct} & \textbf{Multi-step} & \textbf{Direct} & \textbf{Multi-step} \\
        \midrule
        Offensiveness (1-5) & $-0.18$ & $-0.05$ & $+0.15$ & $-0.23$ & $-0.09$ & $-0.94^{*}$ & $-0.38^{*}$ & $-0.38^{*}$ & $+0.12$ & $-0.60^{*}$ \\
        Sarcastic (\% Yes) & $-2.6\%$ & $-2.6\%$ & $+12.5\%^{*}$ & $+2.5\%$ & $+2.9\%$ & $+5.9\%$ & $+5.0\%$ & $+7.5\%$ & $0.0\%$ & $+4.3\%$ \\
        Body Symbol (\% Yes) & $+2.6\%$ & $+2.6\%$ & $+2.5\%$ & $+2.5\%$ & $+2.9\%$ & $+4.7\%$ & $0.0\%$ & $-15.0\%^{*}$ & $+4.8\%$ & $0.0\%$ \\
        Dehumanization (\% Yes) & $-5.1\%$ & $-12.8\%^{*}$ & $-7.5\%$ & $-22.5\%^{*}$ & $-2.9\%$ & $0.0\%$ & $0.0\%$ & $+2.5\%$ & $0.0\%$ & $-2.4\%$ \\
        \midrule
        Sentiment (1-3) & $+0.03$ & $+0.00$ & $+0.10$ & $+0.25$ & $+0.03$ & $+0.24$ & $+0.23$ & $+0.13$ & $+0.12$ & $+0.17$ \\
        Arousal (1-3) & $-0.31$ & $-0.13$ & $-0.20$ & $-0.38$ & $+0.24$ & $-0.15$ & $-0.20$ & $+0.00$ & $+0.10$ & $-0.02$ \\
        Extra Meaning (\% Yes) & $-2.6\%$ & $-20.5\%^{*}$ & $-2.5\%$ & $0.0\%$ & $0.0\%$ & $-2.9\%$ & $+2.5\%$ & $-7.5\%$ & $-2.4\%$ & $0.0\%$ \\
        Clarity (\% Yes) & $-2.6\%$ & $-2.6\%$ & $-10.0\%$ & $0.0\%$ & $-5.9\%$ & $0.0\%$ & $-2.5\%$ & $-2.5\%$ & $+4.8\%$ & $-2.4\%$ \\
        Fluency (\% Yes) & $+5.1\%$ & $+7.7\%$ & $-15.0\%^{*}$ & $-5.0\%$ & $-8.8\%$ & $-8.8\%$ & $+5.0\%$ & $+2.5\%$ & $+9.5\%$ & $+2.4\%$ \\
        \bottomrule
    \end{tabular}%
    }
    \caption{Mean differences in human evaluation metrics after processing tweets using Direct Prompting or our Multi-step Pipeline , relative to the original tweets ($\Delta$ = Processed Score - Original Score), across different offense types. *: $p<0.05$ (paired $t$-test).}
    \label{tab:offense_category_differences}
\end{table*}

\subsection{Heterogeneous Effects by Offensive Types}

While our pipeline reduces overall offensiveness, as discussed in Section \ref{sec:emoji_topics}, users may post offensive tweets of varying types, with emojis serving different functions depending on the nature of the offensiveness. This raises the question of whether our pipeline's effectiveness remains consistent across offense types. We re-calculate the measured variables within each offensive category and the results are presented in Table \ref{tab:offense_category_differences}.

Table \ref{tab:offense_category_differences} shows our multi-step method effectively removed contextually problematic emojis, unlike direct prompting. It reduced dehumanizing symbols by 12.8\% in personal attacks and body-part symbols by 15.0\% in sexual offenses, aligning with the goal of targeting emojis that amplify specific offenses. Corresponding semantic shifts were observed: e.g., emojis conveying `extra meaning' decreased by 20.5\% alongside dehumanizing ones in personal attacks. These aligned changes suggest the pipeline correctly targeted emojis based on their function. Sarcasm levels were generally unaffected, although the direct baseline notably increased perceived sarcasm (+12.5\%) while decreasing fluency (-15.0\%) for political tweets, likely reflecting contextually poor emoji choices.

Crucially, removing problematic emojis did not consistently lower overall offensiveness scores, especially for highly offensive content. Despite reducing dehumanizing emojis in personal (12.8\%) and political (22.5\%) attacks, the change in offensiveness scores for these categories was statistically insignificant. Examining individual cases revealed that offensiveness reduction primarily occurred in mildly offensive tweets (original score < 3). A similar pattern held for sexual offenses: removing body symbols reduced offensiveness in some cases but had no impact on tweets already rated maximally offensive (score=5).
This indicates that while emoji moderation effectively removes specific harmful elements, its impact on overall perceived offensiveness is limited when strong verbal attacks dominate. Emoji replacement appears most effective for mitigating milder forms of offense.

\begin{table}[tbp]
    \resizebox{\linewidth}{!}{%
    \centering
    \small
    \begin{tabular}{p{0.3\linewidth} p{0.3\linewidth} p{0.3\linewidth}}
        \toprule
        \textbf{Case} & \textbf{Original Tweet} & \textbf{Revised Tweet} \\
        \midrule
        No Change in Offensiveness (Score: 4) &  
        @SenateGOP @realDonaldTrump You all are dog doodoo \MyEmoji{\poop} \MyEmoji{\poop}Trump is a disgrace, a traitor and must be removed from office. &  
        @SenateGOP @realDonaldTrump You all are dog doodoo \MyEmoji{\thinkingface}\MyEmoji{\unamusedface} Trump is a disgrace, a traitor and must be removed from office. \\
        \midrule
        Offensiveness Reduced (Score: 2 → 1) &  
        \#TRUMP2020 Landside stable genius \MyEmoji{\poop} &  
        \#TRUMP2020 Landside stable genius \MyEmoji{\unamusedface} \\
        \bottomrule
    \end{tabular}
    }
    \caption{Case study of tweets before and after dehumanizing emoji removal.}
    \label{tab:case_dehumanization}
\end{table}

We present a case study in Table \ref{tab:case_dehumanization}. In the first example, replacing \MyEmoji{\poop} with (\MyEmoji{\thinkingface}, \MyEmoji{\unamusedface}) does not reduce offensiveness, as strong verbal attacks remain. In contrast, the second example shows that removing \MyEmoji{\poop} in a mildly offensive tweet lowers its perceived offensiveness. This suggests that emoji moderation is more effective in less severe cases, while text plays a dominant role in highly offensive tweets.

\section{Implications}
\label{sec:implications}

This work examines emojis' often-overlooked role in online offensiveness. Our analysis of emoji functions and the proposed targeted moderation pipeline offer practical tools and conceptual insights for improving online discourse.
For social media users and platforms, recognizing how emojis subtly shift tone—intensifying, mitigating, or reframing offense—can enhance communication clarity and moderation transparency. It underscores that harm can be conveyed beyond explicit text.

For content moderation researchers and practitioners, our findings emphasize the need to analyze non-textual cues. Emojis implicitly carry offense through symbolism, sarcasm, or stereotypes. We demonstrate that LLMs, guided appropriately like our pipeline, can selectively substitute harmful emojis while preserving semantic content.

For developers of moderation tools, this highlights the importance of incorporating non-verbal signals like emojis, moving beyond current text-centric models. It calls for broader multimodal understanding and opens opportunities for fine-tuning or prompting techniques that address text-emoji interactions in offensive communication.
\section{Conclusion}
\label{sec:focus3_conclusion}

In this paper, we investigate the role of emojis in offensive social media content and propose a multi-step LLM pipeline to mitigate offensiveness while preserving tweet semantics. Our analysis reveals that emojis can amplify, mitigate, or subtly alter offensive content, emphasizing the need for moderation beyond textual cues. Through human evaluation, we demonstrate that our approach effectively reduces offensiveness compared to direct prompting. Our findings highlight the importance of integrating emoji semantics into content moderation and encourage future work to explore adaptive, user-aware moderation strategies.

\section{Limitations}


Despite the promising results, our approach has several key limitations. First, emoji interpretation varies across individuals and cultural backgrounds \cite{lu2016emojiusgae, zhou2024emojis, zhou2024adoption}. In this study, we focus solely on users within the U.S. and use the average offensiveness score to assess the effectiveness of our framework. However, the impact of our pipeline across different linguistic and cultural contexts remains uncertain. A more robust approach would involve cross-linguistic validation and the incorporation of cultural awareness into moderation pipelines.

Second, while we conduct human evaluations to assess offensiveness and semantic preservation before and after applying our pipeline, we lack insights into whether users would accept and adapt to emoji moderation and recommendations in real-world scenarios. User perception and willingness to adjust their emoji usage based on moderation feedback remain unexplored. As a next step, we plan to deploy our system in a user study, collecting feedback through real-time testing and user interviews to better understand the acceptability and usability of emoji moderation.

Third, the effectiveness of our pipeline is inherently tied to the capabilities and potential biases of the LLMs it employs (e.g., GPT-4o, RoBERTa). These models, despite their advancements, can reflect biases present in their training data, potentially leading to skewed interpretations of emoji offensiveness or unfair targeting of certain emoji uses or user expressions. Furthermore, the generalization of the pipeline to novel, rapidly evolving emoji slang or newly introduced emojis is a continuous challenge. LLMs may not immediately grasp the nuanced offensive uses of emojis that emerge after their last training update, requiring ongoing monitoring and model fine-tuning.

\section{Ethical Consideration}
Ethical considerations for the annotation process were carefully observed. The privacy of annotators was protected as no personally identifiable information was collected. The task involved evaluating tweet content and did not entail extensive or intrusive tool usage. In line with our Institutional Review Board's (IRB) protocols for research not involving the collection of identifiable private information about human subjects, this portion of the study was deemed exempt from formal IRB review.

\bibliography{custom}

\appendix

\appendix
\section*{Appendix}

\section{Supplementary Evaluation Materials}

\subsection{Detailed Recruitment Process}

Participant recruitment was conducted via the Prolific online platform, yielding a cohort of 20 annotators. Eligibility for participation was restricted to adults who identified as native English speakers. The annotation protocol involved each individual assessing 60 tweets, a task projected to require approximately 90 minutes of engagement. Compensation was provided at a rate of \$18 USD per hour. All recruited individuals confirmed through responses to initial background inquiries their proficiency and regular interaction with social media environments and emoji use.

\subsection{Detailed Questionnaire of Human Evaluation}

\section*{Annotator Background Questions}
\label{sec:appendix_emoji_annotation}

\begin{enumerate}[label=\arabic*.]
    \item \textbf{What is/are your native language(s)?} \\
    (Please specify in the text box below.) \\
    \textbf{[Fill in the blank]}

    \item \textbf{In the past month, how often do you use social media platforms (e.g., Twitter, Reddit, Instagram, TikTok, Messenger)? Please select the option that best describes the frequency.}
    \begin{itemize}
        \item Never: Never in the past month
        \item Rarely: Fewer than once a week
        \item Sometimes: Two or three times a week
        \item Often: More than three times a week, but not everyday
        \item Always: Almost everyday
    \end{itemize}

    \item \textbf{In the past month, how often do you see content that includes emojis on social media platforms? Please select the option that best describes the frequency.}
    \begin{itemize}
        \item Never: Never in the past month
        \item Rarely: Fewer than once a week
        \item Sometimes: Two or three times a week
        \item Often: More than three times a week, but not everyday
        \item Always: Almost everyday
    \end{itemize}

    \item \textbf{In the past month, how often do you use emojis yourself when engaging on social media platforms? Please select the option that best describes the frequency.}
    \begin{itemize}
        \item Never: Never in the past month
        \item Rarely: Fewer than once a week
        \item Sometimes: Two or three times a week
        \item Often: More than three times a week, but not everyday
        \item Always: Almost everyday
    \end{itemize}
\end{enumerate}

\section*{Guidelines for Assessing Offensiveness}
Offensiveness in the context of tweets refers to content that conveys disrespect, hostility, or harm toward an individual or group. This may occur through negative intent, harmful stereotypes, or inappropriate language. Offensive content is typically characterized by its potential to cause emotional distress or harm to the dignity of those targeted.

When assessing the offensiveness of a tweet, consider the following factors:

\subsection*{Intent Evaluation}
\begin{itemize}
    \item Does the tweet imply a negative or derogatory attitude toward any individual or group?
    \item Assess whether the speaker's intent appears harmful, disrespectful, or mocking.
\end{itemize}

\subsection*{Outcome Evaluation}
\begin{itemize}
    \item Could the tweet cause emotional harm or be perceived as offensive by the targeted individual or group?
    \item Reflect on how the individuals or groups mentioned in the tweet might feel upon reading it.
\end{itemize}

\subsection*{Content Evaluation}
\begin{itemize}
    \item Does the tweet include harmful stereotypes, slurs, or degrading language?
    \item Look for both explicit and implicit elements that might convey disrespect.
\end{itemize}

\subsection*{Acceptability Assessment}
Based on the content and context, determine whether the tweet is socially acceptable:
\begin{itemize}
    \item \textbf{Acceptable}: It is reasonable for the speaker to make this statement.
    \item \textbf{Unacceptable}: The speaker should not have made this statement.
\end{itemize}

\subsection*{Overall Offensiveness Rating}
Rate the tweet’s offensiveness on a scale of 1 to 5:
\begin{itemize}
    \item 1 (Not Offensive At All): No evidence of harmful intent, outcome, or content.
    \item 2 (Slightly Offensive): Minimal or ambiguous evidence of harmful content or intent.
    \item 3 (Moderately Offensive): Clearly problematic in content or intent but without strong emotional harm.
    \item 4 (Highly Offensive): Contains explicit disrespect, harm, or stereotypes; likely to cause emotional harm.
    \item 5 (Extremely Offensive): Extreme disrespect or hostility; causes significant harm or distress.
\end{itemize}

\section*{Examples of Assessing Offensiveness}
To provide more clarity, here are a few examples to illustrate how to evaluate the offensiveness of tweets:

\subsection*{Example 1}
\textbf{Tweet:} "People from [Group X] are lazy and don't want to work."
\begin{itemize}
    \item \textbf{Intent}: Likely derogatory due to stereotyping.
    \item \textbf{Outcome}: Likely to cause emotional harm to Group X.
    \item \textbf{Content}: Contains harmful stereotypes.
    \item \textbf{Acceptability}: Unacceptable.
    \item \textbf{Overall Offensiveness Rating}: 4 (Highly Offensive)
\end{itemize}

\subsection*{Example 2}
\textbf{Tweet:} "I had a terrible experience at [Restaurant Y]."
\begin{itemize}
    \item \textbf{Intent}: Negative but not targeting an individual or group.
    \item \textbf{Outcome}: Unlikely to cause emotional harm.
    \item \textbf{Content}: No harmful language.
    \item \textbf{Acceptability}: Acceptable.
    \item \textbf{Overall Offensiveness Rating}: 1 (Not Offensive At All)
\end{itemize}

\subsection*{Example 3}
\textbf{Tweet:} "This looks like something only a [Group Z] person would wear."
\begin{itemize}
    \item \textbf{Intent}: Ambiguous; may or may not be derogatory.
    \item \textbf{Outcome}: Could cause mild emotional harm to Group Z.
    \item \textbf{Content}: Implicitly reinforces stereotypes about Group Z.
    \item \textbf{Acceptability}: Borderline unacceptable.
    \item \textbf{Overall Offensiveness Rating}: 3 (Moderately Offensive)
\end{itemize}

\section*{Tweet Evaluation Form}

\textbf{Instructions:} Please read the tweet carefully and answer the following questions based on your interpretation. Your responses will help us understand perceptions of the tweet's content.

\paragraph{Tweet:} \textbf{[Display the tweet here]}

\paragraph{Offensiveness Assessment}

\begin{enumerate}[label=\alph*.]
    \item \textbf{Overall Offensiveness} \\
    On a scale from 1 to 5, where 1 means ``Not offensive at all" and 5 means "Extremely offensive," how offensive do you find this tweet?
    \begin{itemize}
        \item 1 - Not offensive at all
        \item 2 - Slightly offensive
        \item 3 - Moderately offensive
        \item 4 - Very offensive
        \item 5 - Extremely offensive
    \end{itemize}

    \item \textbf{Offensive Elements (Optional)} \\
    If you found the tweet offensive, please briefly mention which elements contributed to this perception (e.g., language, emojis, insinuations).
    
    \item \textbf{Is the tweet attempting to be humorous or sarcastic?}
    \begin{itemize}
        \item Yes
        \item No
    \end{itemize}

    \item \textbf{Does the tweet contain any emoji to symbolize human body parts?}
    \begin{itemize}
        \item Yes
        \item No
    \end{itemize}

    \item \textbf{Does the tweet contain any emoji to dehumanize a target?}
    \begin{itemize}
        \item Yes
        \item No
    \end{itemize}
\end{enumerate}

\paragraph{Sentiment Analysis}

\begin{enumerate}[label=\alph*.]
    \item \textbf{Overall Sentiment} \\
    How would you rate the overall sentiment of the tweet?
    \begin{itemize}
        \item 1 - Negative
        \item 2 - Neutral
        \item 3 - Positive
    \end{itemize}

    \item \textbf{Emotional Intensity} \\
    How emotionally intense or stimulating is the tweet? Consider the energy, excitement, or agitation it conveys, regardless of whether the sentiment is positive or negative.
    \begin{itemize}
        \item 1 - Low arousal
        \item 2 - Moderate arousal
        \item 3 - High arousal
    \end{itemize}
\end{enumerate}

\paragraph{Extra Emoji Meaning}

\begin{enumerate}[label=\alph*.]
    \item \textbf{Do you agree with this statement: ``The meanings of all emojis in the tweet are disclosed by the text of the tweet."}
    \begin{itemize}
        \item 1 - Disagree
        \item 2 - Agree
    \end{itemize}
\end{enumerate}

\paragraph{Tweet Attributes}

\begin{enumerate}[label=\alph*.]
    \item \textbf{Clarity} \\
    Is the tweet clearly written? Does this tweet provide enough information for the reader to understand its meaning?
    \begin{itemize}
        \item Yes
        \item No
    \end{itemize}

    \item \textbf{Fluency} \\
    Does the tweet sound fluent or natural? Consider whether the tweet is easy to read and flows smoothly.
    \begin{itemize}
        \item Yes
        \item No
    \end{itemize}
\end{enumerate}

\section{Supplementary Results}
\label{sec:appendix_supplementary_result}

\begin{table*}[tbp]
\centering
\begin{tabular}{l p{9cm}}
\toprule
\textbf{Offensive Type} & \textbf{Topics} \\ \hline
Personal Attacks and Disrespect & 0\_Personal Confrontations and Profanity, 1\_Explicit Content Solicitation, 3\_Offensive Language Usage, 6\_Casual Slang and Swearing, 9\_Offensive Language and Slang Usage, 10\_Casual Profanity Usage \\ \hline
Sexual Content and Gender Issues & 4\_Demeaning Language Toward Women \\ \hline
Racial and Ethnic Offense & 2\_Racial Slur Usage in Conversation, 11\_Racial Discrimination and Stereotyping \\ \hline
Political and Social Issues & 5\_Criticism of Trump's Statements, 13\_Offensive Political and Religious Comments \\ \hline
Violence and Abuse & 7\_Child Abuse Concerns, 8\_Sleep and Fatigue, 12\_Student Violence \\ \bottomrule

\end{tabular}
\caption{\label{table:offense_topic} Categories of offensive types and the belonged topics. Numbers before each topic represents the topic number, ranked by the number of tweets in this topic.}
\end{table*}

\subsection{Qualitative Evaluation: Emoji Distribution Comparison}
\label{sec:appendix_emoji_dist}

We present the emoji distribution for each offensive type before and after running our pipeline in Table \ref{table:emoji_comparison}.

\begin{table}[tbp]
    \resizebox{\linewidth}{!}{%
    \small
    \centering
\begin{tabular}{lll}
\toprule
Offense Type                                     & Pipeline         & Emoji Distribution \\ \midrule
\multirow{3}{*}{Sexual Content}                  & Original         &  \MyEmoji{\droplets}\MyEmoji{\eggplant}\MyEmoji{\tongue}\MyEmoji{\hearteyes}\MyEmoji{\peach}\MyEmoji{\horns}\MyEmoji{\facesavoringfood}\MyEmoji{\droolingface}\MyEmoji{\fire}\MyEmoji{\kissmark}                  \\
& Direct prompting &   \MyEmoji{\hearteyes}\MyEmoji{\droplets}\MyEmoji{\smilingfacesmilingeyes}\MyEmoji{\facesavoringfood}\MyEmoji{\fire}\MyEmoji{\peach}\MyEmoji{\grinningface}\MyEmoji{\eggplant}\MyEmoji{\horns}\MyEmoji{\kissmark}              \\ & Multi-step       &   \MyEmoji{\smilingfacesmilingeyes}\MyEmoji{\grinningface}\MyEmoji{\waveemoji}\MyEmoji{\goldstar}\MyEmoji{\stuckouttonguewinkingface}\MyEmoji{\facesavoringfood}\MyEmoji{\thinkingface}\MyEmoji{\neutralface}\MyEmoji{\hearteyes}\MyEmoji{\cherryblossom}                \\ \midrule
\multirow{3}{*}{Personal Attacks} & Original         &   \MyEmoji{\poop}\MyEmoji{\joy}\MyEmoji{\clownface}\MyEmoji{\facewithrollingeyes}\MyEmoji{\cursingface}\MyEmoji{\vomitingface}\MyEmoji{\ratemoji}\MyEmoji{\rollinglaughing}\MyEmoji{\middlefinger}\MyEmoji{\rage}                 \\
& Direct prompting &   \MyEmoji{\joy}\MyEmoji{\smirk}\MyEmoji{\rage}\MyEmoji{\grinningface}\MyEmoji{\facewithsteam}\MyEmoji{\clownface}\MyEmoji{\thinkingface}\MyEmoji{\angryface}\MyEmoji{\facewithrollingeyes}\MyEmoji{\rollinglaughing}                 \\
& Multi-step       & \MyEmoji{\grinningface}\MyEmoji{\neutralface}\MyEmoji{\unamusedface}\MyEmoji{\thinkingface}\MyEmoji{\confusedface}\MyEmoji{\facewithsteam}\MyEmoji{\angryface}\MyEmoji{\rage}\MyEmoji{\thumbsdown}\MyEmoji{\smilingfacesmilingeyes}                   \\ \midrule
\multirow{3}{*}{Racial Offense}       & Original         &  \MyEmoji{\joy}\MyEmoji{\hundredpoints}\MyEmoji{\sob}\MyEmoji{\woozyface}\MyEmoji{\rollinglaughing}\MyEmoji{\skull}\MyEmoji{\facewithrollingeyes}\MyEmoji{\speakinghead}\MyEmoji{\trashemoji}\MyEmoji{\smirk}                  \\
& Direct prompting &   \MyEmoji{\joy}\MyEmoji{\hundredpoints}\MyEmoji{\sob}\MyEmoji{\grinningface}\MyEmoji{\smirk}\MyEmoji{\neutralface}\MyEmoji{\rage}\MyEmoji{\thinkingface}\MyEmoji{\rollinglaughing}\MyEmoji{\facewithrollingeyes}                 \\
& Multi-step       &  \MyEmoji{\grinningface}\MyEmoji{\neutralface}\MyEmoji{\thinkingface}\MyEmoji{\thumbsup}\MyEmoji{\smilingfacesmilingeyes}\MyEmoji{\unamusedface}\MyEmoji{\confusedface}\MyEmoji{\rage}\MyEmoji{\smirk}\MyEmoji{\grinningsquintingface}                  \\ \midrule
\multirow{3}{*}{Political Issues}     & Original         &    \MyEmoji{\ratemoji}\MyEmoji{\poop}\MyEmoji{\cursingface}\MyEmoji{\rage}\MyEmoji{\clownface}\MyEmoji{\joy}\MyEmoji{\middlefinger}\MyEmoji{\usaflag}                 \\
& Direct prompting &   \MyEmoji{\mouseface}\MyEmoji{\clownface}\MyEmoji{\thinkingface}\MyEmoji{\rage}\MyEmoji{\unamusedface}\MyEmoji{\lyingface}\MyEmoji{\thumbsdown}\MyEmoji{\usaflag}\MyEmoji{\poop}\MyEmoji{\unamusedface}                 \\
& Multi-step       &   \MyEmoji{\thinkingface}\MyEmoji{\neutralface}\MyEmoji{\unamusedface}\MyEmoji{\angryface}\MyEmoji{\clownface}\MyEmoji{\rage}\MyEmoji{\usaflag}\MyEmoji{\grinningface}\MyEmoji{\lyingface}\MyEmoji{\confusedface}                 \\ \midrule
\multirow{3}{*}{Violence and Abuse}              & Original         &  \MyEmoji{\cursingface}\MyEmoji{\rage}\MyEmoji{\facewithsteam}\MyEmoji{\joy}\MyEmoji{\angryface}\MyEmoji{\speakinghead}\MyEmoji{\hundredpoints}\MyEmoji{\skull}\MyEmoji{\sob}\MyEmoji{\unamusedface}                  \\
& Direct prompting &  \MyEmoji{\rage}\MyEmoji{\facewithsteam}\MyEmoji{\unamusedface}\MyEmoji{\grinningface}\MyEmoji{\speakinghead}\MyEmoji{\sob}\MyEmoji{\fire}\MyEmoji{\hundredpoints}\MyEmoji{\joy}\MyEmoji{\pensiveface}                 \\
& Multi-step       &   \MyEmoji{\neutralface}\MyEmoji{\facewithsteam}\MyEmoji{\rage}\MyEmoji{\confusedface}\MyEmoji{\unamusedface}\MyEmoji{\grinningface}\MyEmoji{\prohibitedemoji}\MyEmoji{\unamusedface}\MyEmoji{\thumbsup}\MyEmoji{\pensiveface}                 \\ \bottomrule
\end{tabular}
}
    \caption{Top-10 frequent emojis under different offense types before and after emoji substitution with our proposed pipeline.}
    \label{table:emoji_comparison}
\end{table}

\subsection{LLM Pipeline Prompting Details}
\label{sec:appendix_direct_prompt}

This appendix provides the detailed prompts used in each step of the LLM pipeline described in Section \ref{sec:llm_reduce}.

\paragraph{Step 1: Offensive Content Classification Prompt}

You are tasked with analyzing a tweet for offensive content. Determine if the tweet contains any offensive language or sentiments. Offensive content includes any form of non-acceptable language (profanity) or a targeted offense (veiled or direct), such as insults, threats, profane language, or swear words.

If offensive content is detected, identify the type of offense based on the following categories:
\begin{enumerate}
    \item \textbf{Sexual Content and Gender Issues}: Includes sexual harassment, gender discrimination, body shaming, objectification, gender-based insults, and derogation.
    \item \textbf{Personal Attacks and Disrespect}: Ranges from direct insults to subtle disrespect/derogation targeting individuals/groups based on personal characteristics.
    \item \textbf{Racial and Ethnic Offense}: Includes racial slurs, ethnic stereotyping, and discrimination/prejudice based on race or ethnicity.
    \item \textbf{Political and Social Issues}: Encompasses political attacks, social discrimination, harassment, and aggression over political views or social status.
    \item \textbf{Violence and Abuse}: Includes topics related to physical or verbal abuse, violence, threats, and aggressive behaviors.
\end{enumerate}

\textbf{Your Response Should Include:}
\begin{itemize}
    \item Whether the tweet is offensive (\textit{Yes/No}).
    \item If offensive, the offense category (\textit{e.g., Sexual Content and Gender Issues}).
    \item A brief justification for the chosen category.
\end{itemize}

\textbf{Examples:}
\begin{itemize}
    \item \textbf{Tweet:} \texttt{@sinnersworldxxx \MyEmoji{\rocket}\MyEmoji{\unamusedface}\MyEmoji{\kissmark}\MyEmoji{\growingheart} RT to f*p with a surprise in DM \#sex \#horny https://t.co/VtTjihrWSP}
    \item \textbf{Offensive:} Yes
    \item \textbf{Offense Category:} Sexual Content and Gender Issues
    \item \textbf{Justification:} This tweet promotes sexually suggestive behavior (\textit{e.g., "RT to f*p"}) and references adult content (\textit{e.g., \#sex, \#horny}), violating standards around explicit material.

    \item \textbf{Tweet:} \texttt{My throat hurts. Can god give me a break \MyEmoji{\smilingfacewithhearts}.}
    \item \textbf{Offensive:} No
    \item \textbf{Justification:} This tweet expresses personal discomfort casually and contains no inappropriate or offensive language.
\end{itemize}

\textbf{Tweet to classify:}
\begin{itemize}
    \item \textbf{Tweet:} \promptplaceholder{tweet}
\end{itemize}

\paragraph{Step 2: Emoji Role Determination Prompt}

You will be given an offensive tweet identified as type: \promptplaceholder{offense\_type}. Analyze each emoji (e.g., \MyEmoji{\joy}) within this tweet to determine its functionality. An emoji may fit multiple categories.

The functionality categories are:
\begin{enumerate}
    \item Emoji represents the \textbf{offense itself}
    \item Emoji  \textbf{intensifies} the offense
    \item Emoji  \textbf{mitigates} the offense
    \item Emoji  is \textbf{not directly related} to the offense
\end{enumerate}

Use the following data (emojis based on provided list) to inform your analysis:

\textbf{Emojis Often Associated with Specific Offense Types:}
\begin{itemize}
    \item \textit{Sexual Content:} \MyEmoji{\joy}, \MyEmoji{\rollinglaughing}, \MyEmoji{\sob}, \MyEmoji{\hearteyes}, \MyEmoji{\woozyface}, \MyEmoji{\facesavoringfood}, \MyEmoji{\droolingface}, \MyEmoji{\wearyface}, \MyEmoji{\horns}, \MyEmoji{\facewithrollingeyes}, \MyEmoji{\smirk}, \MyEmoji{\droplets}, \MyEmoji{\pleadingface}, \MyEmoji{\eyes}, \MyEmoji{\tongue}, \MyEmoji{\fire}, \MyEmoji{\thinkingface}, \MyEmoji{\unamusedface}, \MyEmoji{\skull}, \MyEmoji{\heart}, \MyEmoji{\kissmark}, \MyEmoji{\hotface}
    \item \textit{Personal Attacks:} \MyEmoji{\joy}, \MyEmoji{\sob}, \MyEmoji{\rollinglaughing}, \MyEmoji{\facewithrollingeyes}, \MyEmoji{\clownface}, \MyEmoji{\woozyface}, \MyEmoji{\skull}, \MyEmoji{\poop}, \MyEmoji{\vomitingface}, \MyEmoji{\unamusedface}, \MyEmoji{\thinkingface}, \MyEmoji{\rage}, \MyEmoji{\cursingface}, \MyEmoji{\flushedface}, \MyEmoji{\pensiveface}, \MyEmoji{\nauseated}, \MyEmoji{\facewithsteam}, \MyEmoji{\neutralface}, \MyEmoji{\angryface}, \MyEmoji{\expressionlessface}, \MyEmoji{\eyes}, \MyEmoji{\pleadingface}, \MyEmoji{\grinningface}, \MyEmoji{\wearyface}, \MyEmoji{\wink}, \MyEmoji{\speakinghead}
    \item \textit{Racial/Ethnic:} \MyEmoji{\joy}, \MyEmoji{\sob}, \MyEmoji{\rollinglaughing}, \MyEmoji{\woozyface}, \MyEmoji{\unamusedface}, \MyEmoji{\skull}, \MyEmoji{\facewithrollingeyes}, \MyEmoji{\hundredpoints}, \MyEmoji{\speakinghead}, \MyEmoji{\thinkingface}, \MyEmoji{\nauseated}
    \item \textit{Political/Social:} \MyEmoji{\joy}, \MyEmoji{\rage}, \MyEmoji{\facewithrollingeyes}, \MyEmoji{\wink}, \MyEmoji{\nauseated}, \MyEmoji{\facewithopenmouth}, \MyEmoji{\smirk}, \MyEmoji{\sob}
    \item \textit{Violence/Abuse:} \MyEmoji{\joy}, \MyEmoji{\sob}, \MyEmoji{\rollinglaughing}, \MyEmoji{\skull}, \MyEmoji{\rage}, \MyEmoji{\facewithrollingeyes}, \MyEmoji{\nauseated}, \MyEmoji{\hundredpoints}, \MyEmoji{\cursingface}, \MyEmoji{\woozyface}, \MyEmoji{\unamusedface}, \MyEmoji{\hearteyes}, \MyEmoji{\fire}
\end{itemize}

\textbf{Popular Emojis by Typical Functionality:}
\begin{itemize}
    \item \textit{Offense Itself (1):} \MyEmoji{\clownface}, \MyEmoji{\poop}, \MyEmoji{\middlefinger}, \MyEmoji{\joy}, \MyEmoji{\middlefinger}, \MyEmoji{\rage}, \MyEmoji{\rollinglaughing}, \MyEmoji{\middlefinger}, \MyEmoji{\horns}, \MyEmoji{\facewithrollingeyes}, \MyEmoji{\vomitingface}, \MyEmoji{\dizzy}, \MyEmoji{\middlefinger}, \MyEmoji{\ratemoji}, \MyEmoji{\sob}, \MyEmoji{\wearyface}, \MyEmoji{\lyingface}, \MyEmoji{\cursingface}, \MyEmoji{\hundredpoints}
    \item \textit{Intensify Offense (2):} \MyEmoji{\joy}, \MyEmoji{\sob}, \MyEmoji{\rollinglaughing}, \MyEmoji{\facewithrollingeyes}, \MyEmoji{\skull}, \MyEmoji{\woozyface}, \MyEmoji{\wearyface}, \MyEmoji{\unamusedface}, \MyEmoji{\hundredpoints}, \MyEmoji{\hearteyes}, \MyEmoji{\facesavoringfood}, \MyEmoji{\fire}, \MyEmoji{\facewithsteam}, \MyEmoji{\nauseated}, \MyEmoji{\rage}, \MyEmoji{\clownface}, \MyEmoji{\horns}, \MyEmoji{\vomitingface}, \MyEmoji{\droolingface}, \MyEmoji{\cursingface}, \MyEmoji{\thinkingface}, \MyEmoji{\smirk}, \MyEmoji{\droplets}, \MyEmoji{\speakinghead}, \MyEmoji{\hotface}
    \item \textit{Mitigate Offense (3):} \MyEmoji{\joy}, \MyEmoji{\sob}, \MyEmoji{\pleadingface}, \MyEmoji{\pensiveface}, \MyEmoji{\rollinglaughing}, \MyEmoji{\wearyface}, \MyEmoji{\woozyface}, \MyEmoji{\heart}, \MyEmoji{\cryingface}, \MyEmoji{\flushedface}, \MyEmoji{\upsidedownface}, \MyEmoji{\smilingfacewithhearts}, \MyEmoji{\skull}, \MyEmoji{\twohearts}, \MyEmoji{\smilingfacesmilingeyes}, \MyEmoji{\neutralface}, \MyEmoji{\confusedface}, \MyEmoji{\sleepingface}, \MyEmoji{\hearteyes}, \MyEmoji{\relieved}, \MyEmoji{\kiss}, \MyEmoji{\openhands}, \MyEmoji{\stuckouttonguewinkingface}
    \item \textit{Unrelated (4):} \MyEmoji{\joy}, \MyEmoji{\sob}, \MyEmoji{\hearteyes}, \MyEmoji{\rollinglaughing}, \MyEmoji{\thinkingface}, \MyEmoji{\fire}, \MyEmoji{\wearyface}, \MyEmoji{\woozyface}, \MyEmoji{\heart}, \MyEmoji{\eyes}, \MyEmoji{\pleadingface}, \MyEmoji{\skull}, \MyEmoji{\smilingfacewithhearts}, \MyEmoji{\smilingfacewithsunglasses}, \MyEmoji{\flushedface}, \MyEmoji{\speakinghead}, \MyEmoji{\blueheart}, \MyEmoji{\droolingface}, \MyEmoji{\hundredpoints}, \MyEmoji{\blackheart}
\end{itemize}

\textbf{Emoji  Frequency in Offensive Tweets:}
\begin{itemize}
    \item \textit{High Freq.:} \MyEmoji{\middlefinger}, \MyEmoji{\poop}, \MyEmoji{\eggplant}, \MyEmoji{\ratemoji}, \MyEmoji{\cursingface}, \MyEmoji{\trashemoji}
    \item \textit{Medium Freq.:} \MyEmoji{\vomitingface}, \MyEmoji{\tongue}, \MyEmoji{\droplets}, \MyEmoji{\clownface}, \MyEmoji{\nauseated}, \MyEmoji{\angryface}, \MyEmoji{\peach}, \MyEmoji{\rage}, \MyEmoji{\unamusedface}, \MyEmoji{\horns}, \MyEmoji{\facewithsteam}, \MyEmoji{\skull}, \MyEmoji{\speakinghead}
    \item \textit{Low Freq.:} \MyEmoji{\facewithrollingeyes}, \MyEmoji{\woozyface}, \MyEmoji{\rollinglaughing}, \MyEmoji{\joy}, \MyEmoji{\droolingface}, \MyEmoji{\hundredpoints}, \MyEmoji{\facesavoringfood}, \MyEmoji{\hotface}, \MyEmoji{\kissmark}, \MyEmoji{\smirk}, \MyEmoji{\sob}, \MyEmoji{\wearyface}, \MyEmoji{\thinkingface}, \MyEmoji{\fire}, \MyEmoji{\kiss}, \MyEmoji{\hearteyes}
\end{itemize}

\textbf{Your Task:} For each emoji  in the provided tweet:
\begin{itemize}
    \item State the emoji  (e.g., \MyEmoji{\sob}).
    \item Assign functionality category/categories (1-4).
    \item Provide a brief justification based on context and the data provided.
\end{itemize}

\textbf{Examples:}
\begin{itemize}
    \item \textbf{Tweet:} \texttt{Y’all seriously why do her boobs look like analog sticks\MyEmoji{\sob}\MyEmoji{\rollinglaughing}}
    \item \textbf{Emojis:} \MyEmoji{\sob}, \MyEmoji{\rollinglaughing}
    \item \textbf{Analysis for \MyEmoji{\sob}:}
        \begin{itemize}
            \item \textbf{Functionality:} 2 (Intensifies), 4 (Unrelated)
            \item \textbf{Justification:} Amplifies mockery, making the comment seem excessively funny, reinforcing disrespect. Also unrelated to the core offensive content itself.
        \end{itemize}
    \item \textbf{Analysis for \MyEmoji{\rollinglaughing}:}
         \begin{itemize}
            \item \textbf{Functionality:} 2 (Intensifies), 4 (Unrelated)
            \item \textbf{Justification:} Emphasizes the offensive joke as humorous, diminishing seriousness. Also unrelated to the core offensive content.
        \end{itemize}

    \item \textbf{Tweet:} \texttt{just remembered some of the stupid sh*t I did last night, how do I get myself in these situations \MyEmoji{\joy}\MyEmoji{\joy}}
    \item \textbf{Emojis:} \MyEmoji{\joy}, \MyEmoji{\joy}
    \item \textbf{Analysis for \MyEmoji{\joy} (first):}
        \begin{itemize}
            \item \textbf{Functionality:} 3 (Mitigates), 4 (Unrelated)
            \item \textbf{Justification:} Downplays potential offense with humor/self-deprecation, framing it as non-serious.
        \end{itemize}
     \item \textbf{Analysis for \MyEmoji{\joy} (second):}
        \begin{itemize}
            \item \textbf{Functionality:} 3 (Mitigates), 4 (Unrelated)
            \item \textbf{Justification:} Repetition further emphasizes humor, reducing seriousness.
        \end{itemize}

    \item \textbf{Tweet:} \texttt{F**k de calor\MyEmoji{\hotface}}
    \item \textbf{Emojis:} \MyEmoji{\hotface}
    \item \textbf{Analysis for \MyEmoji{\hotface}:}
        \begin{itemize}
            \item \textbf{Functionality:} 2 (Intensifies)
            \item \textbf{Justification:} Intensifies the expressed frustration/discomfort, aligning with the profanity to amplify sentiment intensity.
        \end{itemize}

    \item \textbf{Tweet:} \texttt{Like and RT for more clip\MyEmoji{\droplets}\#cum [...] \#nuses… https://t.co/gSNIDclo6s} 
    \item \textbf{Emojis:} \MyEmoji{\droplets} 
     \item \textbf{Analysis for \MyEmoji{\droplets}:}
        \begin{itemize}
            \item \textbf{Functionality:} 1 (Offense Itself), 2 (Intensifies)
            \item \textbf{Justification:} Often implies sexual acts/arousal, directly contributing to and intensifying the explicit content.
        \end{itemize}
\end{itemize}

\textbf{Tweet to analyze:}
\begin{itemize}
    \item \textbf{Tweet:} \promptplaceholder{tweet}
    \item \textbf{Emojis in Tweet:} \promptplaceholder{emoji}
\end{itemize}

\paragraph{Step 3: Emoji Surrogate Recommendation Prompt}

You are given an offensive tweet (type: \promptplaceholder{offense\_type}) where emojis have been classified by functionality:
\begin{enumerate}
    \item Emoji  represents the \textbf{offense itself}
    \item Emoji  \textbf{intensifies} the offense
    \item Emoji  \textbf{mitigates} the offense
    \item Emoji  is \textbf{not directly related} to the offense
\end{enumerate}

Your task is to recommend replacements only for emojis categorized as \textbf{1 (Offense Itself)} or \textbf{2 (Intensifies Offense)}. Emojis categorized as \textbf{3 (Mitigates)} or \textbf{4 (Unrelated)} should be kept.

For emojis needing replacement:
\begin{itemize}
    \item Suggest a replacement emoji  (e.g., \MyEmoji{\thinkingface}) that maintains the tweet's general tone but mitigates the specific offense.
    \item Choose replacements consistent with the tweet's original content context.
\end{itemize}

\textbf{Your Response Should Include (for each replaced emoji):}
\begin{itemize}
    \item Emojis to Replace: {Emoji}
    \item Replacement Emojis: {Emoji}
    \item Justification: Brief explanation for the replacement choice.
    \item (Finally) Revised Tweet: The full tweet after all necessary replacements.
\end{itemize}

\textbf{Examples:}
\begin{itemize}
    \item \textbf{Tweet:} \texttt{Y’all seriously why do her boobs look like analog sticks\MyEmoji{\sob}\MyEmoji{\rollinglaughing}}
    \item \textbf{Emojis/Functionality:} \MyEmoji{\sob} (2, 4), \MyEmoji{\rollinglaughing} (2, 4)
    \item \textit{Replacement 1:}
        \begin{itemize}
            \item Emoji to Replace: \MyEmoji{\sob}
            \item Replacement Emoji: \MyEmoji{\thinkingface}
            \item Justification: Maintains playful tone but shifts from ridicule to curiosity/confusion.
        \end{itemize}
    \item \textit{Replacement 2:}
         \begin{itemize}
            \item Emoji to Replace: \MyEmoji{\rollinglaughing}
            \item Replacement Emoji: \MyEmoji{\smilingfacesmilingeyes}
            \item Justification: Keeps tweet lighthearted without amplifying offense; signals amusement softly.
        \end{itemize}
    \item \textbf{Revised Tweet:} \texttt{Y’all seriously why do her boobs look like analog sticks\MyEmoji{\thinkingface}\MyEmoji{\smilingfacesmilingeyes}}

    \item \textbf{Tweet:} \texttt{Omg yes! \MyEmoji{\fistraised}\MyEmoji{\wearyface} sex havers roll call \MyEmoji{\horns}} 
    \item \textbf{Emojis/Functionality:} \MyEmoji{\fistraised} (4), \MyEmoji{\wearyface} (2), 
     \MyEmoji{\horns} (1, 2)
    \item \textit{Replacement 1:}
        \begin{itemize}
            \item Emojis to Replace: \MyEmoji{\wearyface}
            \item Replacement Emojis: \MyEmoji{\grinningface}
            \item Justification: Retains excitement but softens intensity; conveys playful enthusiasm without explicit connotations.
        \end{itemize}
    \item \textit{Replacement 2:}
        \begin{itemize}
            \item Emojis to Replace: \MyEmoji{\horns}
            \item Replacement Emojis: \MyEmoji{\stuckouttonguewinkingface}
            \item Justification: Maintains mischievous tone but mitigates offense; more lighthearted/less suggestive.
        \end{itemize}
    \item \textbf{Revised Tweet:} \texttt{Omg yes! \MyEmoji{\fistraised}\MyEmoji{\grinningface} sex havers roll call \MyEmoji{\stuckouttonguewinkingface}} 
\end{itemize}

\textbf{Tweet to process:}
\begin{itemize}
    \item \textbf{Tweet:} \promptplaceholder{tweet}
    \item \textbf{Emojis in Tweet (with functionalities):} \promptplaceholder{emoji}
\end{itemize}

\paragraph{Step 4: Justification Summary Generation Prompt}

You will be given an original offensive tweet, its revised version where some emoji  were replaced to mitigate offense, and justifications for the offense type, original emoji functionalities, and emoji replacements.

Your task is to **summarize** these justifications into a single, concise paragraph explaining \textit{why} specific emoji  were replaced.

Recall the emoji functionality categories:
\begin{enumerate}
    \item Offense Itself
    \item Intensifies Offense
    \item Mitigates Offense
    \item Unrelated to Offense
\end{enumerate}
We only replaced emojis categorized as \textbf{1} or \textbf{2}.

\textbf{Your Response Should Include:}
\begin{itemize}
    \item A summary paragraph integrating the offense type, the functionality of the replaced emojis, and the reason for their replacements.
\end{itemize}

\textbf{Inputs Provided:}
\begin{itemize}
    \item Original Tweet: \promptplaceholder{tweet}
    \item Emojis in Original Tweet: \promptplaceholder{emoji}
    \item Revised Tweet: \promptplaceholder{revised\_tweet}
    \item Justification of Tweet Offense Type: \promptplaceholder{offense\_type}
    \item Justification of Emoji Functionality: \promptplaceholder{emoji\_func}
    \item Justification of Emoji Replacement: \promptplaceholder{emoji\_replace}
\end{itemize}

\textbf{Generate Justification Summary:}

\paragraph{Prompt of Direct Prompting}
The prompt for the direct prompting baseline is: \\ \textit{You will be given a tweet with emojis. If this tweet is offensive, try to only replace the emojis with ones that maintain the tweet's tone but mitigate the offense. If the tweet is non-offensive, provide the original tweet as the revised tweet.}\\

\subsection{Qualitative Evaluation: Case Studies}
\label{sec:appendix_case_study}

\begin{figure*}[tbp]
     \centering
     \includegraphics[width=\linewidth]{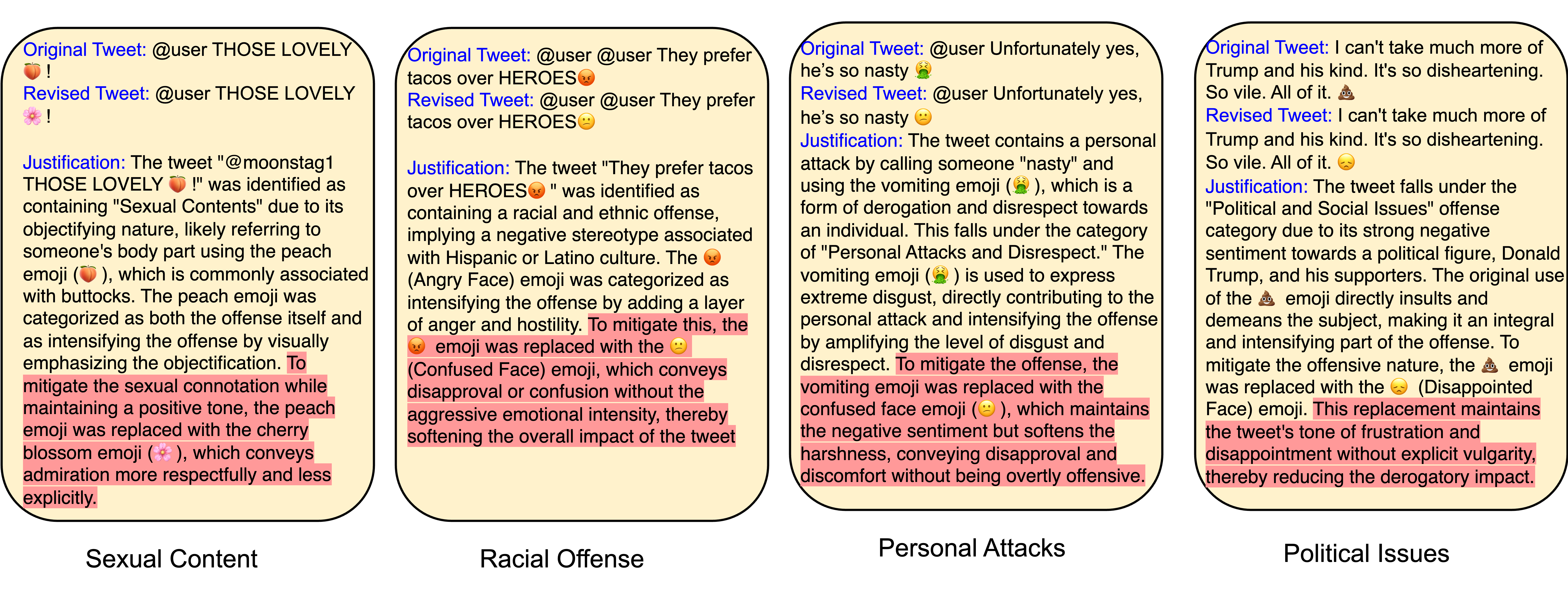}
     \caption{\label{fig:case_study_pipeline} Justification from our multi-step pipeline of emoji replacement. The red color highlights the reason of choosing that emoji surrogate. The offense type of each tweet is labeled below.}
\end{figure*}

We present four random examples of the original tweet and the revised tweet after processing through our pipeline, covering different offensive types in Figure \ref{fig:case_study_pipeline}. In addition, we include the justifications summarized by the LLMs in the final step of our pipeline for each emoji substitution.

The examples and justifications presented in Figure \ref{fig:case_study_pipeline} demonstrate that our pipeline effectively identifies offensive content, provides the reasoning behind its offensiveness, and captures the role of emojis within the tweet. For instance, in the first example, the LLM accurately interprets \MyEmoji{\peach} as a reference to a body part, which intensifies the offense. It suggests replacing it with the flower emoji \MyEmoji{\cherryblossom} to keep the positive sentiment while reducing the offensive nature of the tweet. 
Moreover, in the second example, where the post includes the word ``tacos'' and the \MyEmoji{\rage} emoji, our pipeline detects the implicit racial offense toward Hispanic or Latino culture. It recommends replacing \MyEmoji{\rage} with \MyEmoji{\confusedface} to reduce the offense while maintaining the semantics of the post.
In conclusion, our pipeline effectively identifies offensive content within tweets, uncovers the relationship between emojis and offensive material, and precisely recommends emoji surrogates to mitigate the offense while preserving the tweet's overall semantics.

While our case study demonstrates the effectiveness of our multi-step pipeline, the next step is to quantitatively assess its impact on offensiveness reduction for each tweet. In the following section, we leverage human annotations to evaluate the offensiveness of tweets before and after LLM rewriting.

\end{document}